\documentclass[10pt,twocolumn,letterpaper]{article}

\usepackage[pagenumbers]{cvpr} 
\usepackage{tabularx}
\usepackage[table]{xcolor}
\usepackage{booktabs}
\usepackage{array}
\usepackage{multirow}
\usepackage[table,dvipsnames]{xcolor}
\usepackage{booktabs}    
\definecolor{vdarkblue}{RGB}{0,32,91}      
\definecolor{darkblue}{RGB}{33,102,172}     
\definecolor{medblue}{RGB}{103,169,207}     
\definecolor{lightblue}{RGB}{171,217,233}   
\definecolor{verylightblue}{RGB}{209,229,240} 
\definecolor{almostwhite}{RGB}{247,247,247} 
\definecolor{verylightbeige}{RGB}{253,219,199} 
\definecolor{lightbeige}{RGB}{253,187,132}  
\definecolor{medbeige}{RGB}{252,141,89}     
\definecolor{orange}{RGB}{227,74,51}        
\definecolor{darkred}{RGB}{179,0,0}  

\usepackage{graphicx}
\usepackage{float}

\definecolor{cvprblue}{rgb}{0.21,0.49,0.74}
\usepackage[pagebackref,breaklinks,colorlinks,allcolors=cvprblue]{hyperref}

\def\paperID{***}
\def\confName{CVPR}
\def\confYear{2026}

\title{Can Vision-Language Models Count? A Synthetic Benchmark and Analysis of Attention-Based Interventions}

\author{Saurav Sengupta \thanks{These authors contributed equally.}
\and Nazanin Moradinasab $^*$
\and Jiebei Liu $^*$
\and Donald E. Brown\\
School of Data Science, University of Virginia\\
{\tt\small {\{ss4yd, nm4wu, mcu2xn, deb\}}@virginia.edu}
}

\begin{document}
\maketitle
\begin{abstract}
Recent research suggests that Vision Language Models (VLMs) often rely on inherent biases learned during training when responding to queries about visual properties of images. These biases are exacerbated when VLMs are asked highly specific questions that require selective visual attention, a demand that mirrors cognitive challenges observed in human enumeration tasks. We build upon this research by developing a synthetic benchmark dataset and evaluation framework to systematically characterize how counting performance varies as image and prompt properties change. Using open-source VLMs, we analyze how performance shifts across controlled perturbations (e.g. number of objects, object color, background color, object texture, background texture, and prompt specificity) and examine corresponding changes in visual attention allocation. We further conduct exploratory attention reweighting experiments in the language model decoder to modulate focus on visual tokens at different layers and assess their effects on counting behavior. Our results reveal that counting accuracy degrades systematically with increasing visual and linguistic complexity echoing human limits and cognitive load effects known from human perception, while targeted attention reweighting yields modest but measurable improvements. Rather than competing on benchmark accuracy, we introduce a controlled diagnostic framework for analyzing VLM enumeration behavior. Through systematic experiments, we expose failure modes rooted in cross-modal binding that natural image benchmarks may not easily isolate, and provide preliminary empirical evidence that targeted attention reweighting in the language decoder can influence how models ground linguistic quantity concepts in visual representations. Code and data available here: \url{https://github.com/ssen7/vlm-count-analysis}

\end{abstract}   
\vspace{-0.25cm}
\section{Introduction}
\label{sec:intro}

Specialized counting methods consistently outperform VLMs on controlled benchmarks, including SAM-based frameworks like PseCo \cite{huang2024point}, open-world counters like CountGD that use grounding and visual exemplars \cite{amini2024countgd}, and diffusion-based density estimators like CrowdDiff \cite{ranasinghe2024crowddiff}. Yet despite this gap, robust counting remains an essential capability for VLMs. Unlike specialized counting tools that are trained for specific object types, VLMs need to handle open-ended questions where counting is just one part of a broader task. From both cognitive and architectural perspectives, accurate estimation of object quantities can serve as a foundational capability for general visual reasoning and downstream applications in robotics, medical imaging and more \cite{shao2025large, ryu2025vision}. We systematically evaluate how current open-source VLMs perform on counting tasks under varying prompt styles and image conditions, and analyze how attention patterns are associated with performance changes

Prior work shows that VLMs struggle with counting \cite{guo2025your,alghisi2025re}. 
Our work differs from existing approaches by introducing a fine-grained diagnostic framework that analyzes VLM counting behavior across four dimensions: (1) prompt specificity, (2) visual complexity (texture, color, background), (3) object counts ranges  (0–50 in increments of 10), and (4) attention distribution over vision tokens. This controlled setup allows us to correlate specific visual and linguistic inputs (1–3) with their mechanistic effects on model internals (4), enabling a more precise understanding of both counting capabilities and failure modes.
We evaluate three open-source VLMs that perform strongly on established benchmarks: \textbf{Qwen-VL} \cite{qwen2.5-VL}, \textbf{Kimi-VL-A3B} \cite{kimiteam2025kimivltechnicalreport}, and \textbf{InternVL3-9B} \cite{chen2024internvl}. These models represent complementary architectural and inference paradigms, including dense transformer architecture and Mixture-of-Experts (MoE) design. For selected models, we evaluate dedicated reasoning and instruction-tuned checkpoints, allowing us to isolate whether explicit chain-of-thought inference confers any advantage in counting tasks. Their open-source nature is central to our approach, as it enables direct introspection into model internals. 

We further use a subset of our evaluation data to investigate how targeted modifications to attention over vision tokens affect counting performance. This is motivated in part by recent work showing that VLMs exhibit a systematic tendency to concentrate high attention weights on specific visual tokens that are largely irrelevant to the input query, called visual attention sinks \cite{kang2025see}.
Our primary evaluation uses synthetic images by design as synthetic stimuli allow controlled manipulation of factors that are typically entangled in real-world data, such as occlusion, texture, density, and background clutter. To assess whether our findings generalize, we validate selected interventions on a subset of the FSC-147 real-world counting benchmark \cite{ranjan2021learning}. Despite the increased visual complexity of natural scenes, we observe similar qualitative trends, suggesting that the identified failure modes are not artifacts of synthetic stimuli.

In summary, in this paper we:

\begin{enumerate}
    \item Create a synthetic counting dataset that allows us to evaluate VLMs counting performance. The images, while unchallenging for modern computer vision, allow us to isolate specific characteristics of the image and control for other variables to investigate which characteristics of the input (prompt/image) can affect counting performance.
    \item Characterize the chosen VLM's counting performance along a) increasing prompt specificity, b) object shapes and colors, c) background colors and textures, and d) distribution of attention over vision tokens.
    \item We investigate how attention interventions over visual tokens in the language decoder,both globally and at individual layers, shape model counting behavior, probing whether redirecting attention toward object-bearing regions leads to measurable improvements in accuracy.
\end{enumerate}

\section{Related Work}
\label{sec:related_work}
Recent research has revealed significant limitations in Vision Language Models' counting capabilities.  Vo et al.\citep{vlmsarebiased} demonstrate that state-of-the-art VLMs like Open AI's o3 \cite{o3} and Google DeepMind's Gemini 2.5 Pro \cite{gemini} exhibit strong prior knowledge biases that severely compromise counting accuracy, as models default to memorized patterns rather than analyzing visual features.  This bias extends to general visual understanding.  Guo et al. \cite{guo2025your} systematically show that VLMs struggle to count beyond 20 objects, with performance deteriorating as scene complexity increases. Aghisi at al. \cite{alghisi2025re}  identify that reasoning chains can partially improve counting, with middle transformer layers being most critical for accurate enumeration. 

Attention mechanisms play a crucial role in VLM visual processing. Kang et al. \cite{kang2025see} reveal that VLMs often allocate excessive attention to irrelevant sections of the image and propose a method to redistribute the attention to more relevant areas. An et al. \cite{an2025mitigating} demonstrate that object hallucinations can be mitigated through specialized attention mechanisms that better integrate global and local visual features.  
Several studies have examined the vision understanding of VLMs by evaluating their Visual Question Answering (VQA) capabilities using synthetic imaging data \cite{hou2024vision, lee2024vlind, lee2024vhelm} or by visualizing Attention Guided Class Activation Maps (AG-CAM) on chart-based data \cite{dong2025probing}. 

Our work builds upon these foundations by providing a more granular, diagnostic benchmark. We systematically isolate the impact of \textit{visual properties} (e.g., textures, colors, shapes) to pinpoint failure modes related to visual complexity, while also varying input prompts along a specificity gradient, from generic to increasingly precise descriptions, to disentangle the contributions of linguistic and visual factors. 
To probe the attention mechanisms underlying counting failures, we implement two complementary intervention strategies: image-naive layer-wise modifications that systematically suppress or amplify attention across individual decoder layers, and mask-guided interventions that incorporate spatial object information to more precisely redirect attention toward object-bearing regions. Together, these interventions allow us to evaluate both the sources of counting failure and the conditions under which targeted attention modifications can improve accuracy.
\vspace{-0.25cm}
\section{Methodology}
\label{sec:method}

\subsection{Synthetic Evaluation Dataset}
We create a collection of synthetic datasets, each consisting of images and corresponding prompts, to facilitate a systematic examination of VLMs.  Each dataset is designed to evaluate distinct aspects of the input data—both visual and textual. 
Our generation process begins with a \textbf{baseline dataset} consisting of 512×512 pixel images containing non-overlapping black circular objects on a pure white background. We generate 50 images for this initial configuration. From this baseline, we iterate by varying two primary factors: 1. \textbf{Object Numerosity:} We vary the object counts in buckets of 10, ensuring each bucket is equally represented. This allows us to evaluate how VLM performance degrades as the number of objects successively increases. 2. \textbf{Visual Properties:} We employ a controlled variable methodology. While preserving the object locations from the base dataset, we systematically vary only one feature at a time from the set:  \{\texttt{object shape}, \texttt{object color}, \texttt{object texture}, \texttt{background color}, \texttt{background texture}\}. All other variables are held constant. 
This systematic generation framework allows us to measure and visualize variations in model performance and shifts in attention allocation. By manipulating a single dimension at a time, we can isolate the specific effects of each visual or textual property on the model's counting capabilities. Refer to Supplementary Section \ref{sec:sample_images_supp}.



\subsection{Evaluation Criteria}
We evaluate VLM counting performance using two primary metrics: Accuracy and Mean Relative Count Error (MRCE). MRCE is defined as:
\begin{equation}
    \text{MRCE} = \frac{1}{N} \sum_{i=1}^{N} \frac{|c_{\text{pred}}^{(i)} - c_{\text{true}}^{(i)}|}{c_{\text{true}}^{(i)}}
\end{equation}
where $N$ is the number of samples, $c_{\text{pred}}^{(i)}$ is the predicted count for sample $i$, and $c_{\text{true}}^{(i)}$ is the ground truth count for sample $i$.

We assess these metrics across four primary experimental axes:

\textbf{Increasing prompt specificity.}
We analyze model sensitivity to prompt phrasing using a ``prompt ladder," where we progressively add descriptive details (e.g., color, texture) to a generic counting prompt. 
Example prompts are provided in Table \ref{tab:texture_prompt_ladder_compact}.

\begin{table}[!h]
\centering
\small 
\caption{Example prompts used when image has different Object Texture.}
\label{tab:texture_prompt_ladder_compact}
\begin{tabularx}{\columnwidth}{l >{\raggedright\arraybackslash}p{0.45\columnwidth} >{\raggedright\arraybackslash}p{0.45\columnwidth}}
\toprule
\textbf{ID} & \textbf{Example Prompt Text} & \textbf{Logical Role / Cognitive Cue} \\
\midrule
P1 & \texttt{Count the number of distinct objects in this image...} & \textbf{Baseline:} Generic unconstrained prompt.   \\
\addlinespace
P2 & \texttt{Count the number of \{color\} color objects in this image...} & \textbf{Single (Simple) Attribute:} Simple Cue (Color) - Replace \{color\} with object colors (``Blue-green" for default). \\
\addlinespace
P3 & \texttt{Count the number of objects with \{pattern\} pattern in this image ...} & \textbf{Single (Complex) Attribute:} Complex  (Texture). Replace \{pattern\} with ``dots", ``linear gradient", ``checkerboard",``vertical stripe", etc.\\
\addlinespace
P4 & \texttt{Count the number of \{pattern\} pattern with \{color\} color objects in this image...} & \textbf{Compositional (Target):}  Binding (Complex + Simple). Tests binding a simple cue with a complex one. \\
\addlinespace
P5 & \texttt{Count the number of \{pattern\} pattern with \{color\} color \{shape\} in this image...} & \textbf{Compositional (High Load):} Multi-attribute binding under high cognitive load.  \\
\bottomrule
\end{tabularx}
\end{table}

\textbf{Sensitivity to Visual Properties. }We evaluate performance by systematically varying a single image characteristic at a time (e.g., \texttt{object color}, \texttt{object texture}, \texttt{background color}, \texttt{background texture}) while holding all other factors constant allowing us to isolate the impact of specific visual features on counting performance. 

\textbf{Object Counts Ranges.} We analyze how VLM performance varies across different object count intervals (e.g., 0–9, 10–19, …, 40–50) to identify the threshold at which a model's counting capabilities begin to deteriorate.  

\textbf{Attention over vision tokens.}
We quantify how much attention is given to visual tokens in each of the above scenarios to glean insight as to how self-attention is choosing to allocate attention over objects in the image. We generate heatmaps using Layer-wise propagation of Visual Attention (LPV) \cite{Chefer_2021_ICCV} and GradCAM \cite{selvaraju2020grad} over all decoder layers and calculate overlap between high attention areas and the objects in the image (See Supplementary Section \ref{sec:lpv}).

\subsection{Impact of attention redistribution over counting performance}

VLMs exhibit systematic failures in counting tasks, with performance degrading sharply as object count increases (See Section \ref{sec:results:levels}). We hypothesize that this limitation stems from diffuse attention mechanisms that fail to distinguish and distinctly represent each object instance, leading to feature conflation and under-counting.

We systematically modify how much attention each layer of the language model decoder pays to each token in the vision input, measuring the resulting impact on accuracy and MRCE. We investigate five attention reweighting strategies that manipulate how VLMs allocate attention between visual and textual tokens during generation. Let $\mathbf{A} \in \mathbb{R}^{H \times Q \times K}$ denote the attention weight matrix for a given layer, where $H$ is the number of attention heads, $Q$ is the query sequence length, and $K$ is the key sequence length. We denote the visual token positions as $V = \{v_{\text{start}}, \ldots, v_{\text{end}}\}$ where visual tokens typically occupy the prefix of the sequence.

\textbf{Amplify.} This strategy increases attention weights to visual tokens by a multiplicative factor $\alpha > 1$:
\begin{equation}
\tilde{A}_{h,i,j} = \begin{cases}
\alpha \cdot A_{h,i,j} & \text{if } j \in V \\
A_{h,i,j} & \text{otherwise}
\end{cases}
\end{equation}
followed by renormalization: $\tilde{A}_{h,i,:} \leftarrow \tilde{A}_{h,i,:} / \sum_{k} \tilde{A}_{h,i,k}$. 
This strategy strengthens visual grounding by encouraging stronger connections to image features. We use $\alpha = 2.0$.

\textbf{Suppress.} Conversely, this strategy reduces attention to visual tokens, forcing greater reliance on linguistic context :
\begin{equation}
\tilde{A}_{h,i,j} = \begin{cases}
\beta \cdot A_{h,i,j} & \text{if } j \in V \\
A_{h,i,j} & \text{otherwise}
\end{cases}
\end{equation}
where $0 < \beta < 1$, followed by renormalization. We set $\beta = 0.5$. 

\textbf{Focus.} This strategy creates an extreme form of visual attention by largely eliminating attention to non-visual tokens:
\begin{equation}
\tilde{A}_{h,i,j} = \begin{cases}
A_{h,i,j} & \text{if } j \in V \\
\epsilon & \text{otherwise}
\end{cases}
\end{equation}
where $\epsilon = 10^{-10}$ is a small constant to maintain numerical stability. After renormalization, attention is effectively concentrated solely on visual tokens, forcing direct visual conditioning at each generation step.

\textbf{Balance.} This strategy enforces a target distribution between visual and textual attention. Given the desired visual attention ratio $r_v^{\text{target}}$  (we use $r_v^{\text{target}} = 0.4$) and the current ratio: $
r_v^{\text{current}} = \frac{\sum_{j \in V} A_{h,i,j}}{\sum_{k} A_{h,i,k}}$, we apply a corrective scaling:

\begin{equation}
\tilde{A}_{h,i,j} = \begin{cases}
\gamma \cdot A_{h,i,j} & \text{if } j \in V \\
A_{h,i,j} & \text{otherwise}
\end{cases}
\end{equation}
where $\gamma = r_v^{\text{target}} / r_v^{\text{current}}$, followed by renormalization. This preserves visual–textual attention balance and prevents over- or under-reliance on visual information.

\textbf{Visual Mask Amplify.}
Let $\mathbf{M} \in \{0,1\}^{H_{\text{img}} \times W_{\text{img}}}$ denote a binary object mask obtained from an off-the-shelf segmentation model (e.g., SAM~\cite{kirillov2023segment}). For a vision transformer with patch size $p$, we partition the image into a grid of $N_h \times N_w$ patches where $N_h = H_{\text{img}}/p$ and $N_w = W_{\text{img}}/p$.

For each visual token $v_i$ corresponding to patch coordinates $(r, c)$, we compute the \textbf{object overlap ratio}:
\begin{equation}
\rho_i = \frac{1}{p^2} \sum_{x=rp}^{(r+1)p-1} \sum_{y=cp}^{(c+1)p-1} \mathbf{M}(x, y)
\end{equation}
which measures the fraction of the patch covered by object regions. We define the set of object-relevant tokens as:
\begin{equation}
V_{\text{obj}} = \{v_i \in V : \rho_i > \tau\}
\end{equation}
where $\tau$ is an overlap threshold (we use $\tau = 0.1$ to capture patches with at least 10\% object coverage).
The visual mask amplify strategy then applies selective amplification:
\begin{equation}
\tilde{A}_{h,i,j} = \begin{cases}
\alpha_{\text{obj}} \cdot A_{h,i,j} & \text{if } j \in V_{\text{obj}} \\
\alpha_{\text{bg}} \cdot A_{h,i,j} & \text{if } j \in V \setminus V_{\text{obj}} \\
A_{h,i,j} & \text{otherwise}
\end{cases}
\end{equation}
followed by renormalization. We test $\alpha_{\text{obj}} = 2.0$ to strongly emphasize objects and $\alpha_{\text{bg}} = 0.5$ to suppress background, creating a high-contrast attention distribution that prioritizes semantically meaningful content. We also test an ablation of this strategy without background suppression.
In mask-based variants, object masks are obtained using an off-the-shelf segmentation model (SAM3 \cite{carion2025sam}). The masks serve as spatial priors to distinguish object regions from background regions when reweighting visual tokens. 
Together, these strategies enable controlled manipulation of the visual–textual attention trade-off, allowing systematic analysis of how attention allocation influences counting behavior.

\section{Results}
\label{sec:results}

\subsection{Effects of Prompt Specificity}
Results for the effects of increasing linguistic details in the text prompt are shown in Table \ref{tab:prompt-acc-mre-optimized}.

\vspace{-0.2cm}
\begin{table}[h]
\centering
\scriptsize
\definecolor{basered}{RGB}{178,24,43}     
\definecolor{gray0}{RGB}{240,243,240}     
\newcommand{\R}[2]{\cellcolor{basered!#1}{\ifnum#1>40\color{white}\fi#2}}
\newcommand{\BL}[2]{\cellcolor{blue!#1}{\ifnum#1>40\color{white}\fi#2}}
\newcommand{\G}[1]{\cellcolor{gray0}{#1}}
\setlength{\tabcolsep}{1.5pt}
\renewcommand{\arraystretch}{1.06}
\caption{Effect of prompt specificity on counting accuracy and MRCE. Prompt 1 (gray cells) serves as the baseline. For Accuracy: darker red indicates greater improvements, darker blue indicates larger drops. For MRCE: darker red indicates greater error reduction  (better), darker blue indicates increased relative error (worse). ``Bg'' = \textit{Background},  ``Obj'' = \textit{Object}.}

\begin{tabular}{@{}lllcccccccc@{}}
\toprule
Cat. & Feat. & Prompts & \multicolumn{2}{c}{Qwen32b} & \multicolumn{2}{c}{Qwen7b} & \multicolumn{2}{c}{InternVL} & \multicolumn{2}{c}{Kimi} \\
 & & & Acc & MRCE & Acc & MRCE & Acc & MRCE & Acc & MRCE \\
\midrule
Bg & color & P1& \G{0.22} & \G{0.133} & \G{0.223} & \G{0.242} & \G{0.167} & \G{0.133} & \G{0.247} & \G{0.162} \\
  & & P2& \R{20}{+0.047} & \R{30}{-0.032} & \R{20}{+0.013} & \R{50}{-0.087} & \BL{10}{-0.010} & \BL{20}{+0.026} & \R{20}{+0.020} & \R{50}{-0.079} \\
  & & P3& \R{30}{+0.033} & \R{40}{-0.038} & \BL{10}{-0.003} & \R{40}{-0.071} & \R{10}{+0.013} & \BL{20}{+0.029} & \R{30}{+0.033} & \R{50}{-0.086} \\
\midrule
Bg & texture & P1& \G{0.182} & \G{0.282} & \G{0.09} & \G{0.638} & \G{0.213} & \G{0.227} & \G{0.169} & \G{0.452} \\
    & & P2& \BL{10}{-0.003} & \R{30}{-0.091} & \R{70}{+0.078} & \R{100}{-0.433} & \R{10}{+0.009} & \R{30}{-0.060} & \R{80}{+0.095} & \R{100}{-0.355} \\
    & & P3& \BL{20}{-0.018} & \R{30}{-0.075} & \R{70}{+0.078} & \R{100}{-0.415} & \BL{20}{-0.020} & \BL{30}{+0.048} & \R{70}{+0.076} & \R{100}{-0.346} \\
    & & P4& \R{40}{+0.042} & \R{60}{-0.149} & \R{60}{+0.062} & \R{100}{-0.409} & \BL{20}{-0.013} & \BL{20}{+0.029} & \R{60}{+0.069} & \R{100}{-0.346} \\
    & & P5& \R{50}{+0.057} & \R{60}{-0.152} & \R{50}{+0.057} & \R{100}{-0.400} & \BL{10}{-0.011} & \BL{10}{+0.020} & \R{60}{+0.067} & \R{100}{-0.350} \\
\midrule
Obj & color & P1& \G{0.24} & \G{0.104} & \G{0.163} & \G{0.274} & \G{0.22} & \G{0.100} & \G{0.246} & \G{0.129} \\
    & & P2& \BL{30}{-0.023} & \BL{20}{+0.021} & \R{50}{+0.049} & \R{60}{-0.115} & \BL{10}{-0.011} & \BL{20}{+0.030} & \BL{10}{-0.003} & \R{10}{-0.013} \\
    & & P3& \BL{40}{-0.043} & \BL{10}{+0.002} & \R{50}{+0.051} & \R{60}{-0.118} & \BL{30}{-0.031} & \BL{20}{+0.034} & \G{0.000} & \R{10}{-0.007} \\
\midrule
Obj& shape & P1& \G{0.196} & \G{0.135} & \G{0.18} & \G{0.347} & \G{0.224} & \G{0.128} & \G{0.228} & \G{0.139} \\
   & & P2& \R{10}{+0.004} & \R{10}{-0.011} & \R{20}{+0.020} & \R{100}{-0.213} & \R{20}{+0.020} & \BL{10}{+0.003} & \R{20}{+0.016} & \BL{50}{+0.090} \\
   & & P3& \R{20}{+0.024} & \BL{10}{+0.002} & \R{10}{+0.004} & \R{100}{-0.203} & \R{20}{+0.024} & \BL{10}{+0.019} & \R{30}{+0.028} & \R{10}{-0.018} \\
\midrule
Obj & texture & P1& \G{0.24} & \G{0.145} & \G{0.172} & \G{0.543} & \G{0.254} & \G{0.113} & \G{0.272} & \G{0.076} \\
    & & P2& \BL{80}{-0.084} & \BL{70}{+0.132} & \BL{10}{-0.006} & \R{100}{-0.319} & \G{0.000} & \BL{30}{+0.048}& \BL{10}{-0.010} & \BL{40}{+0.067}\\
    & & P3& \BL{90}{-0.088} & \BL{60}{+0.107} & \BL{70}{-0.078} & \R{80}{-0.266} & \BL{100}{-0.136} & \BL{100}{+0.309} & \BL{20}{-0.018} & \BL{20}{+0.027}\\
    & & P4& \BL{90}{-0.096} & \BL{80}{+0.144} & \BL{60}{-0.060} & \R{80}{-0.238} & \BL{70}{-0.080} & \BL{90}{+0.201} & \BL{40}{-0.040} & \BL{50}{+0.090}\\
    & & P5& \BL{100}{-0.108} & \BL{90}{+0.172} & \BL{70}{-0.076} & \R{70}{-0.219} & \BL{70}{-0.076} & \BL{90}{+0.194} & \BL{40}{-0.042} & \BL{60}{+0.098} \\
\bottomrule
\end{tabular}
\label{tab:prompt-acc-mre-optimized}
\end{table}

Across all models, prompt specificity has a strongly asymmetric effect depending on the feature type. For background features, specificity consistently improves performance—background texture yields the largest gains, with Qwen7b and Kimi reducing MRCE by 0.433 and 0.355, respectively at P2, persisting through P5. In contrast, object texture is the only category where specificity monotonically degrades accuracy across all models (Qwen32b $\Delta Acc = -0.108$ at P5), even as MRCE improves for some models, suggesting errors become more systematic rather than random.
Object color and shape show mixed, model-dependent responses. Qwen7b benefits substantially from shape specificity ($\Delta MRCE=-0.213$ at P2), while Kimi degrades under the same condition, and InternVL remains largely neutral, indicating greater robustness to prompt variation. Notably, model scale does not confer robustness: Qwen32b degrades substantially on object texture despite being the largest model. Together, these results suggest that specificity aids counting when it simplifies visual segmentation (background cues), but becomes detrimental when it introduces competing object-level processing demands.
\begin{figure*}[h]
    \centering
    \includegraphics[width=0.7\textwidth]{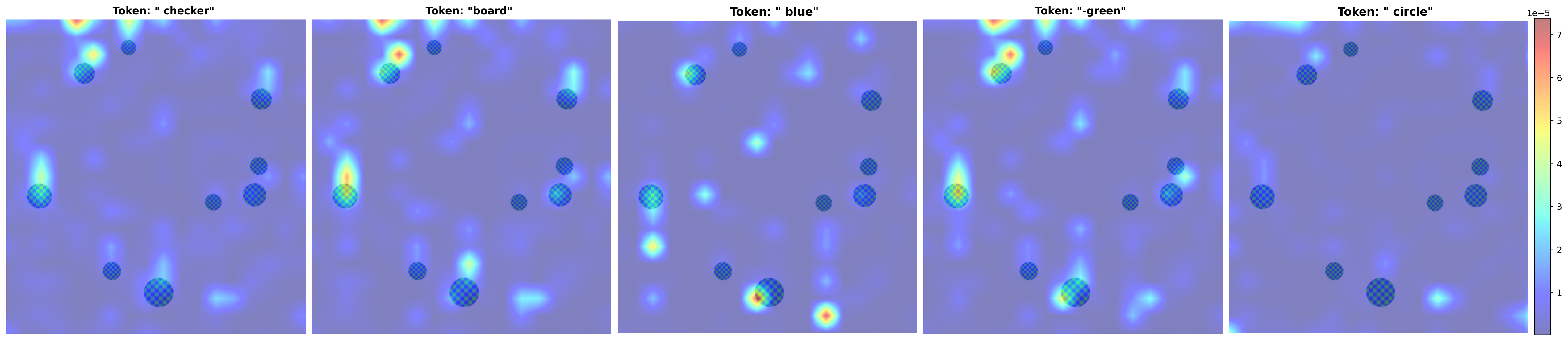}
    \caption{Token-level attention heatmaps for the compositional prompt 5 of object texture task for Kimi. The high load from texture and color suppresses attention to shape}
    \label{fig:att_heatmap_texture_prompt5}
\end{figure*}

P1 (the simplest, most general prompt) succeeds because its very generality allows the model to deploy its most robust internal detector, bypassing the ``cognitive sink" that \textit{any} specific semantic cue (whether \texttt{texture}, \texttt{color}, or \texttt{shape}) creates in this task. We have direct visual evidence for this ``sink" in Figure \ref{fig:att_heatmap_texture_prompt5} when \texttt{shape} is added in P5 (\texttt{color}+\texttt{texture}+\texttt{shape}), attention overlays confirm the model's attention to the object's shape is absent, suppressed by the cognitive load of processing \texttt{texture} and \texttt{color}. 

\subsection{Effects of Visual Complexity } 
\begin{table}[h]
\caption{Mean Relative Count Error (lower is better) for Prompt 2 across all patterns. 
}
\label{tab:error_texture}
\renewcommand{\arraystretch}{0.85} 
\setlength{\tabcolsep}{3pt}        
\centering

\begin{tabular}{@{}llp{1.5cm}cccc@{}}
\toprule
Cat. & Feat. & Pattern & Qwen7b & Qwen32b & Intern & Kimi \\
\midrule
Bg & Color & blue & \cellcolor{blue!55}{\textcolor{black}{0.129}} & \cellcolor{blue!58}{\textcolor{black}{0.109}} & \cellcolor{blue!46}{\textcolor{black}{0.134}} & \cellcolor{blue!57}{\textcolor{black}{0.075}} \\
Bg & Color & black & \cellcolor{blue!53}{\textcolor{black}{0.152}} & \cellcolor{blue!59}{\textcolor{black}{0.088}} & \cellcolor{blue!46}{\textcolor{black}{0.135}} & \cellcolor{blue!57}{\textcolor{black}{0.083}} \\
Bg & Color & green & \cellcolor{blue!52}{\textcolor{black}{0.159}} & \cellcolor{blue!58}{\textcolor{black}{0.109}} & \cellcolor{blue!46}{\textcolor{black}{0.134}} & \cellcolor{blue!59}{\textcolor{black}{0.062}} \\
Bg & Color & gray & \cellcolor{blue!52}{\textcolor{black}{0.156}} & \cellcolor{blue!58}{\textcolor{black}{0.106}} & \cellcolor{blue!40}{\textcolor{black}{0.160}} & \cellcolor{blue!58}{\textcolor{black}{0.072}} \\
Bg & Color & red & \cellcolor{blue!53}{\textcolor{black}{0.150}} & \cellcolor{blue!58}{\textcolor{black}{0.100}} & \cellcolor{blue!39}{\textcolor{black}{0.165}} & \cellcolor{blue!56}{\textcolor{black}{0.087}} \\
Bg & Color & yellow & \cellcolor{blue!50}{\textcolor{black}{0.176}} & \cellcolor{blue!59}{\textcolor{black}{0.095}} & \cellcolor{blue!28}{\textcolor{black}{0.221}} & \cellcolor{blue!54}{\textcolor{black}{0.113}}
\\
\hline
Bg & Texture & lin. grad. & \cellcolor{blue!54}{\textcolor{black}{0.138}} & \cellcolor{blue!58}{\textcolor{black}{0.105}} & \cellcolor{blue!57}{\textcolor{black}{0.077}} & \cellcolor{blue!57}{\textcolor{black}{0.078}} \\
Bg & Texture & noise & \cellcolor{blue!56}{\textcolor{black}{0.120}} & \cellcolor{blue!59}{\textcolor{black}{0.096}} & \cellcolor{blue!43}{\textcolor{black}{0.145}} & \cellcolor{blue!58}{\textcolor{black}{0.071}} \\
Bg & Texture & rad. grad. & \cellcolor{blue!53}{\textcolor{black}{0.146}} & \cellcolor{blue!53}{\textcolor{black}{0.164}} & \cellcolor{blue!48}{\textcolor{black}{0.121}} & \cellcolor{blue!59}{\textcolor{black}{0.062}} \\
Bg & Texture & cr. hatch & \cellcolor{blue!51}{\textcolor{black}{0.172}} & \cellcolor{blue!55}{\textcolor{black}{0.137}} & \cellcolor{blue!46}{\textcolor{black}{0.133}} & \cellcolor{blue!60}{\textcolor{black}{0.053}} \\
Bg & Texture & ver. str. & \cellcolor{blue!47}{\textcolor{black}{0.210}} & \cellcolor{blue!56}{\textcolor{black}{0.130}} & \cellcolor{blue!37}{\textcolor{black}{0.179}} & \cellcolor{blue!59}{\textcolor{black}{0.062}} \\
Bg & Texture & checkerboard & \cellcolor{blue!44}{\textcolor{black}{0.232}} & \cellcolor{blue!55}{\textcolor{black}{0.135}} & \cellcolor{blue!35}{\textcolor{black}{0.188}} & \cellcolor{blue!54}{\textcolor{black}{0.113}} \\
Bg & Texture & hor. str. & \cellcolor{blue!43}{\textcolor{black}{0.245}} & \cellcolor{blue!51}{\textcolor{black}{0.182}} & \cellcolor{blue!50}{\textcolor{black}{0.115}} & \cellcolor{blue!50}{\textcolor{black}{0.154}} \\
Bg & Texture & con. rgs & \cellcolor{blue!44}{\textcolor{black}{0.239}} & \cellcolor{blue!50}{\textcolor{black}{0.193}} & \cellcolor{blue!22}{\textcolor{black}{0.253}} & \cellcolor{blue!58}{\textcolor{black}{0.068}} \\
Bg & Texture & diag. str & \cellcolor{blue!42}{\textcolor{black}{0.249}} & \cellcolor{blue!41}{\textcolor{black}{0.308}} & \cellcolor{blue!43}{\textcolor{black}{0.147}} & \cellcolor{blue!58}{\textcolor{black}{0.071}} \\
Bg & Texture & bubbles & \cellcolor{blue!41}{\textcolor{black}{0.264}} & \cellcolor{blue!52}{\textcolor{black}{0.176}} & \cellcolor{blue!12}{\textcolor{black}{0.300}} & \cellcolor{blue!45}{\textcolor{black}{0.204}} \\
Bg & Texture & dots & \cellcolor{blue!41}{\textcolor{black}{0.264}} & \cellcolor{blue!20}{\textcolor{black}{0.560}} & \cellcolor{blue!38}{\textcolor{black}{0.174}} & \cellcolor{blue!54}{\textcolor{black}{0.116}}
\\
\hline
Obj & Color & green & \cellcolor{blue!59}{\textcolor{black}{0.097}} & \cellcolor{blue!60}{\textcolor{black}{0.086}} & \cellcolor{blue!56}{\textcolor{black}{0.085}} & \cellcolor{blue!58}{\textcolor{black}{0.066}} \\
Obj & Color & red & \cellcolor{blue!56}{\textcolor{black}{0.126}} & \cellcolor{blue!57}{\textcolor{black}{0.115}} & \cellcolor{blue!55}{\textcolor{black}{0.088}} & \cellcolor{blue!59}{\textcolor{black}{0.063}} \\
Obj & Color & blue & \cellcolor{blue!57}{\textcolor{black}{0.108}} & \cellcolor{blue!58}{\textcolor{black}{0.105}} & \cellcolor{blue!49}{\textcolor{black}{0.120}} & \cellcolor{blue!59}{\textcolor{black}{0.062}} \\
Obj & Color & white & \cellcolor{blue!58}{\textcolor{black}{0.099}} & \cellcolor{blue!58}{\textcolor{black}{0.101}} & \cellcolor{blue!52}{\textcolor{black}{0.102}} & \cellcolor{blue!54}{\textcolor{black}{0.109}} \\
Obj & Color & yellow & \cellcolor{blue!54}{\textcolor{black}{0.141}} & \cellcolor{blue!59}{\textcolor{black}{0.095}} & \cellcolor{blue!53}{\textcolor{black}{0.100}} & \cellcolor{blue!54}{\textcolor{black}{0.114}} \\
Obj & Color & light gray & \cellcolor{blue!51}{\textcolor{black}{0.170}} & \cellcolor{blue!57}{\textcolor{black}{0.113}} & \cellcolor{blue!35}{\textcolor{black}{0.187}} & \cellcolor{blue!57}{\textcolor{black}{0.079}} \\
Obj & Color & multicolor & \cellcolor{blue!30}{\textcolor{black}{0.362}} & \cellcolor{blue!45}{\textcolor{black}{0.259}} & \cellcolor{blue!28}{\textcolor{black}{0.223}} & \cellcolor{blue!36}{\textcolor{black}{0.308}} \\
\hline
Obj & Shape & star & \cellcolor{blue!55}{\textcolor{black}{0.130}} & \cellcolor{blue!57}{\textcolor{black}{0.117}} & \cellcolor{blue!57}{\textcolor{black}{0.077}} & \cellcolor{blue!57}{\textcolor{black}{0.076}} \\
Obj & Shape & polygon & \cellcolor{blue!57}{\textcolor{black}{0.110}} & \cellcolor{blue!58}{\textcolor{black}{0.105}} & \cellcolor{blue!45}{\textcolor{black}{0.140}} & \cellcolor{blue!56}{\textcolor{black}{0.089}} \\
Obj & Shape & rectangle & \cellcolor{blue!60}{\textcolor{black}{0.089}} & \cellcolor{blue!56}{\textcolor{black}{0.128}} & \cellcolor{blue!33}{\textcolor{black}{0.198}} & \cellcolor{blue!57}{\textcolor{black}{0.079}} \\
Obj & Shape & circle & \cellcolor{blue!51}{\textcolor{black}{0.168}} & \cellcolor{blue!56}{\textcolor{black}{0.131}} & \cellcolor{blue!38}{\textcolor{black}{0.173}} & \cellcolor{blue!58}{\textcolor{black}{0.073}} \\
Obj & Shape & triangle & \cellcolor{blue!52}{\textcolor{black}{0.161}} & \cellcolor{blue!55}{\textcolor{black}{0.137}} & \cellcolor{blue!60}{\textcolor{black}{0.066}} & \cellcolor{blue!31}{\textcolor{black}{0.360}} \\
\hline
Obj & Texture & lin. grad. & \cellcolor{blue!53}{\textcolor{black}{0.147}} & \cellcolor{blue!58}{\textcolor{black}{0.105}} & \cellcolor{blue!55}{\textcolor{black}{0.089}} & \cellcolor{blue!58}{\textcolor{black}{0.064}} \\
Obj & Texture & con. cir. & \cellcolor{blue!55}{\textcolor{black}{0.134}} & \cellcolor{blue!58}{\textcolor{black}{0.104}} & \cellcolor{blue!49}{\textcolor{black}{0.116}} & \cellcolor{blue!58}{\textcolor{black}{0.074}} \\
Obj & Texture & rad. grad. & \cellcolor{blue!54}{\textcolor{black}{0.137}} & \cellcolor{blue!53}{\textcolor{black}{0.164}} & \cellcolor{blue!54}{\textcolor{black}{0.095}} & \cellcolor{blue!58}{\textcolor{black}{0.068}} \\
Obj & Texture & ver. str. & \cellcolor{blue!56}{\textcolor{black}{0.124}} & \cellcolor{blue!56}{\textcolor{black}{0.130}} & \cellcolor{blue!48}{\textcolor{black}{0.125}} & \cellcolor{blue!56}{\textcolor{black}{0.095}} \\
Obj & Texture & checkerboard & \cellcolor{blue!49}{\textcolor{black}{0.191}} & \cellcolor{blue!49}{\textcolor{black}{0.206}} & \cellcolor{blue!50}{\textcolor{black}{0.113}} & \cellcolor{blue!58}{\textcolor{black}{0.072}} \\
Obj & Texture & hor. str. & \cellcolor{blue!51}{\textcolor{black}{0.167}} & \cellcolor{blue!51}{\textcolor{black}{0.182}} & \cellcolor{blue!37}{\textcolor{black}{0.175}} & \cellcolor{blue!57}{\textcolor{black}{0.079}} \\
Obj & Texture & zigzag & \cellcolor{blue!51}{\textcolor{black}{0.173}} & \cellcolor{blue!49}{\textcolor{black}{0.207}} & \cellcolor{blue!33}{\textcolor{black}{0.199}} & \cellcolor{blue!58}{\textcolor{black}{0.073}} \\
Obj & Texture & diag. str. & \cellcolor{blue!40}{\textcolor{black}{0.270}} & \cellcolor{blue!41}{\textcolor{black}{0.308}} & \cellcolor{blue!42}{\textcolor{black}{0.150}} & \cellcolor{blue!59}{\textcolor{black}{0.060}} \\
Obj & Texture & dots & \cellcolor{blue!44}{\textcolor{black}{0.232}} & \cellcolor{blue!20}{\textcolor{black}{0.560}} & \cellcolor{blue!37}{\textcolor{black}{0.177}} & \cellcolor{blue!52}{\textcolor{black}{0.139}} \\
Obj & Texture & cr. hatch & \cellcolor{blue!10}{\textcolor{black}{0.652}} & \cellcolor{blue!10}{\textcolor{black}{0.797}} & \cellcolor{blue!10}{\textcolor{black}{0.362}} & \cellcolor{blue!10}{\textcolor{black}{0.700}} \\
\bottomrule
\end{tabular}
\end{table}


Table~\ref{tab:error_texture} presents the Mean Relative Count Error (MRCE) for Prompt 2—the single-attribute, color-focused prompt— across all background and object variations. The results show a clear trend: model performance degrades substantially as visual complexity increases. Models perform well under simple conditions, such as solid-colored backgrounds and plain, single-color objects, where MRCE is lowest. However, the error increases significantly when high-frequency textures (e.g., checkerboard, diagonal/vertical stripes, concentric rings) or visually heterogeneous objects (multicolor or complex patterns) are introduced. This pattern is consistent across most models, indicating that complex textures and patterns interfere with the models' ability to reliably segment and enumerate objects. 
Results for Prompts 1 and 3–5 are provided in the Supplementary.

\subsection{Effects of Count Magnitude}
\label{sec:results:levels}

To analyze the effect of count magnitude on model accuracy, we divided all images into five discrete count bins as shown in Table~\ref{tab:error_blue_shaded}. 
The results are collapsed across all prompt formulations, where darker blue shades indicate smaller errors.
We observe a consistent trend across all models: counting performance degrades monotonically with increasing object count. In the low-count regime ($<$10), most models exhibit minimal error ($<$0.1), indicating strong reliability when few instances are present. However, as the number of objects increases, error grows non-linearly—particularly beyond 30.
When separating by visual feature type,simple Color category shows the smallest deviation. In contrast, texture categories consistently exhibit the largest variance, implying that complex surface patterns interfere with spatial grouping mechanisms. Among the models, Kimi-VL-A3B-Instruct and Qwen2.5-32B-Instruct maintain the lowest relative errors overall, whereas Qwen2.5-7B-Instruct and InternVL3-9B-Instruct exhibit greater sensitivity under large-count conditions.
These results demonstrate a level-dependent counting robustness: performance remains stable for small sets but declines as visual density and textural complexity increase, emphasizing the need for count-adaptive attention strategies in future architectures.


\begin{table}[!t]
\caption{Mean Relative Count Error (lower is better). Darker blue cells indicate smaller errors. 
}
\label{tab:error_blue_shaded}
\renewcommand{\arraystretch}{0.85} 
\setlength{\tabcolsep}{3pt}        
\centering
\begin{tabular}{@{}lllcccc@{}}     
\toprule
Cat. & Feat.  & Counts & Qwen7b & Qwen32b & Intern & Kimi \\
\midrule
Bg & Color & $<$10 & \cellcolor{blue!60}{\textcolor{white}{0.039}} & \cellcolor{blue!60}{\textcolor{white}{0.050}} & \cellcolor{blue!51}{\textcolor{white}{0.099}} & \cellcolor{blue!60}{\textcolor{white}{0.020}} \\
Bg & Color & 10–19 & \cellcolor{blue!56}{\textcolor{white}{0.068}} & \cellcolor{blue!56}{\textcolor{white}{0.067}} & \cellcolor{blue!53}{\textcolor{white}{0.082}} & \cellcolor{blue!41}{\textcolor{white}{0.085}} \\
Bg & Color & 20–29 & \cellcolor{blue!46}{\textcolor{white}{0.154}} & \cellcolor{blue!40}{\textcolor{white}{0.138}} & \cellcolor{blue!45}{\textcolor{white}{0.131}} & \cellcolor{blue!37}{\textcolor{black}{0.100}} \\
Bg & Color & 30–39 & \cellcolor{blue!37}{\textcolor{black}{0.226}} & \cellcolor{blue!43}{\textcolor{white}{0.124}} & \cellcolor{blue!36}{\textcolor{black}{0.189}} & \cellcolor{blue!30}{\textcolor{black}{0.125}} \\
Bg & Color & 40–50 & \cellcolor{blue!10}{\textcolor{black}{0.451}} & \cellcolor{blue!33}{\textcolor{black}{0.170}} & \cellcolor{blue!25}{\textcolor{black}{0.253}} & \cellcolor{blue!10}{\textcolor{black}{0.201}} \\
\hline
Bg & Texture & $<$10 & \cellcolor{blue!48}{\textcolor{white}{0.137}} & \cellcolor{blue!36}{\textcolor{black}{0.155}} & \cellcolor{blue!45}{\textcolor{white}{0.134}} & \cellcolor{blue!43}{\textcolor{white}{0.077}} \\
Bg & Texture & 10–19 & \cellcolor{blue!46}{\textcolor{white}{0.150}} & \cellcolor{blue!34}{\textcolor{black}{0.167}} & \cellcolor{blue!48}{\textcolor{white}{0.117}} & \cellcolor{blue!25}{\textcolor{black}{0.143}} \\
Bg & Texture & 20–29 & \cellcolor{blue!26}{\textcolor{black}{0.315}} & \cellcolor{blue!26}{\textcolor{black}{0.202}} & \cellcolor{blue!38}{\textcolor{black}{0.176}} & \cellcolor{blue!14}{\textcolor{black}{0.182}} \\
Bg & Texture & 30–39 & \cellcolor{blue!14}{\textcolor{black}{0.411}} & \cellcolor{blue!29}{\textcolor{black}{0.188}} & \cellcolor{blue!30}{\textcolor{black}{0.226}} & \cellcolor{blue!10}{\textcolor{black}{0.232}} \\
Bg & Texture & 40–50 & \cellcolor{blue!10}{\textcolor{black}{0.533}} & \cellcolor{blue!14}{\textcolor{black}{0.255}} & \cellcolor{blue!14}{\textcolor{black}{0.323}} & \cellcolor{blue!10}{\textcolor{black}{0.222}} \\
\hline
Obj & Color & $<$10 & \cellcolor{blue!55}{\textcolor{white}{0.080}} & \cellcolor{blue!54}{\textcolor{white}{0.076}} & \cellcolor{blue!53}{\textcolor{white}{0.081}} & \cellcolor{blue!53}{\textcolor{white}{0.044}} \\
Obj & Color & 10–19 & \cellcolor{blue!52}{\textcolor{white}{0.097}} & \cellcolor{blue!48}{\textcolor{white}{0.103}} & \cellcolor{blue!54}{\textcolor{white}{0.078}} & \cellcolor{blue!40}{\textcolor{white}{0.090}} \\
Obj & Color & 20–29 & \cellcolor{blue!42}{\textcolor{white}{0.182}} & \cellcolor{blue!45}{\textcolor{white}{0.116}} & \cellcolor{blue!49}{\textcolor{white}{0.109}} & \cellcolor{blue!35}{\textcolor{black}{0.107}} \\
Obj & Color & 30–39 & \cellcolor{blue!37}{\textcolor{black}{0.223}} & \cellcolor{blue!39}{\textcolor{black}{0.144}} & \cellcolor{blue!42}{\textcolor{white}{0.150}} & \cellcolor{blue!16}{\textcolor{black}{0.172}} \\
Obj & Color & 40–50 & \cellcolor{blue!16}{\textcolor{black}{0.395}} & \cellcolor{blue!39}{\textcolor{black}{0.142}} & \cellcolor{blue!36}{\textcolor{black}{0.186}} & \cellcolor{blue!11}{\textcolor{black}{0.190}} \\
\hline
Obj & Shape & $<$10 & \cellcolor{blue!54}{\textcolor{white}{0.087}} & \cellcolor{blue!53}{\textcolor{white}{0.079}} & \cellcolor{blue!60}{\textcolor{white}{0.044}} & \cellcolor{blue!56}{\textcolor{white}{0.032}} \\
Obj & Shape & 10–19 & \cellcolor{blue!53}{\textcolor{white}{0.094}} & \cellcolor{blue!46}{\textcolor{white}{0.112}} & \cellcolor{blue!54}{\textcolor{white}{0.079}} & \cellcolor{blue!38}{\textcolor{black}{0.097}} \\
Obj & Shape & 20–29 & \cellcolor{blue!45}{\textcolor{white}{0.156}} & \cellcolor{blue!30}{\textcolor{black}{0.185}} & \cellcolor{blue!39}{\textcolor{black}{0.168}} & \cellcolor{blue!27}{\textcolor{black}{0.135}} \\
Obj & Shape & 30–39 & \cellcolor{blue!30}{\textcolor{black}{0.282}} & \cellcolor{blue!37}{\textcolor{black}{0.151}} & \cellcolor{blue!40}{\textcolor{white}{0.164}} & \cellcolor{blue!16}{\textcolor{black}{0.174}} \\
Obj & Shape & 40–50 & \cellcolor{blue!16}{\textcolor{black}{0.395}} & \cellcolor{blue!41}{\textcolor{white}{0.132}} & \cellcolor{blue!30}{\textcolor{black}{0.223}} & \cellcolor{blue!10}{\textcolor{black}{0.220}} \\
\hline
Obj & Texture & $<$10 & \cellcolor{blue!46}{\textcolor{white}{0.152}} & \cellcolor{blue!27}{\textcolor{black}{0.199}} & \cellcolor{blue!53}{\textcolor{white}{0.085}} & \cellcolor{blue!45}{\textcolor{white}{0.073}} \\
Obj & Texture & 10–19 & \cellcolor{blue!39}{\textcolor{black}{0.210}} & \cellcolor{blue!18}{\textcolor{black}{0.237}} & \cellcolor{blue!37}{\textcolor{black}{0.183}} & \cellcolor{blue!28}{\textcolor{black}{0.130}} \\
Obj & Texture & 20–29 & \cellcolor{blue!20}{\textcolor{black}{0.368}} & \cellcolor{blue!10}{\textcolor{black}{0.273}} & \cellcolor{blue!19}{\textcolor{black}{0.294}} & \cellcolor{blue!26}{\textcolor{black}{0.139}} \\
Obj & Texture & 30–39 & \cellcolor{blue!13}{\textcolor{black}{0.422}} & \cellcolor{blue!16}{\textcolor{black}{0.247}} & \cellcolor{blue!11}{\textcolor{black}{0.341}} & \cellcolor{blue!21}{\textcolor{black}{0.156}} \\
Obj & Texture & 40–50 & \cellcolor{blue!10}{\textcolor{black}{0.515}} & \cellcolor{blue!10}{\textcolor{black}{0.323}} & \cellcolor{blue!10}{\textcolor{black}{0.412}} & \cellcolor{blue!19}{\textcolor{black}{0.162}} \\
\bottomrule
\end{tabular}
\end{table}

\subsection{Reasoning vs Instruction Tuned Models}
We additionally evaluate reasoning-enabled variants of two open-source reasoning VLMs, Qwen3b-Thinking and Kimi-VL-A3B-Thinking models, and compare them to instruction-tuned counterparts. The results show in Table \ref{tab:thinking_vs_instruct}. As shown, reasoning does not consistently improve counting performance. While reasoning models can reduce error for low object counts in some cases, they frequently produce unparsable or verbose outputs at higher counts, leading to reduced effective coverage. For example, Qwen3-Thinking often fails to produce valid numerical answers in dense scenes, and Kimi-Instruct generally outperforms Kimi-Thinking overall. These results suggest that enumeration errors stem primarily from visual grounding limitations rather than insufficient linguistic reasoning.

\begin{table}[!ht]
\caption{
\vspace{-2pt}
Mean Relative Error (MRE) computed over parsable outputs. For Qwen-Thinking, images with $>10$ objects resulted in interminable reasoning loops, while Kimi-Thinking performed noticably worse than Kimi-Instruct.
\vspace{-6pt}
}
\label{tab:thinking_vs_instruct}
\footnotesize

\renewcommand{\arraystretch}{0.60}
\setlength{\tabcolsep}{1.0pt}
\setlength{\aboverulesep}{0pt}
\setlength{\belowrulesep}{0pt}
\setlength{\extrarowheight}{0pt}

\centering
\vspace{-4pt}

{\bfseries Qwen}\par
\begin{tabular}{@{}lllcccc@{}}
\toprule
Cat.&Feat.&Counts&Qwen3-T&Pars.&N-par.&Qwen7b\\
\midrule
Bg&Color&$<$10  &\cellcolor{blue!70}{\textcolor{white}{0.017}}&173&7   &\cellcolor{blue!65}\textcolor{white}{{0.039}}\\
Bg&Color&10--50 &\cellcolor{blue!44}{\textcolor{black}{0.114}}&85 &635 &--\\
\hline
Bg&Txtr &$<$10  &\cellcolor{blue!59}{\textcolor{white}{0.058}}&495&55  &\cellcolor{blue!44}{\textcolor{black}{0.137}}\\
Bg&Txtr &10--50 &\cellcolor{blue!25}{\textcolor{black}{0.182}}&275&1925&--\\
\hline
Obj&Color&$<$10  &\cellcolor{blue!67}{\textcolor{white}{0.028}}&204&6   &\cellcolor{blue!51}{\textcolor{white}{0.080}}\\
Obj&Color&10--50 &\cellcolor{blue!51}{\textcolor{black}{0.086}}&108&732 &--\\
\hline
Obj&Shape&$<$10  &\cellcolor{blue!69}{\textcolor{white}{0.019}}&147&0   &\cellcolor{blue!51}{\textcolor{white}{0.087}}\\
Obj&Shape&10--50 &\cellcolor{blue!44}{\textcolor{black}{0.113}}&86 &517 &--\\
\hline
Obj&Txtr &$<$10  &\cellcolor{blue!61}{\textcolor{white}{0.051}}&479&21  &\cellcolor{blue!33}{0.152}\\
Obj&Txtr &10--50 &\cellcolor{blue!10}{\textcolor{black}{0.239}}&269&1731&--\\
\bottomrule
\end{tabular}

\vspace{1pt} 

{\bfseries Kimi}\par\vspace{0pt}
\begin{tabular}{@{}lllcccc@{}}
\toprule
Cat.&Feat.&Counts&Kimi-T&Pars.&N-par.&Kimi\\
\midrule
Bg&Color&$<$10
&\cellcolor{blue!66}{\textcolor{white}{0.135}}&150&0
&\cellcolor{blue!78}{\textcolor{white}{0.020}}\\
Bg&Color&10--50
&\cellcolor{blue!31}{\textcolor{black}{0.424}}&534&66
&\cellcolor{blue!55}{\textcolor{black}{0.128}}\\
\hline
Bg&Txtr &$<$10
&\cellcolor{blue!43}{\textcolor{black}{0.336}}&587&3
&\cellcolor{blue!63}{\textcolor{white}{0.077}}\\
Bg&Txtr &10--50
&\cellcolor{blue!10}{\textcolor{black}{0.525}}&2358&42
&\cellcolor{blue!46}{\textcolor{black}{0.195}}\\
\hline
Obj&Color&$<$10
&\cellcolor{blue!68}{\textcolor{white}{0.130}}&210&0
&\cellcolor{blue!70}{\textcolor{white}{0.044}}\\
Obj&Color&10--50
&\cellcolor{blue!31}{\textcolor{black}{0.423}}&823&17
&\cellcolor{blue!52}{\textcolor{black}{0.145}}\\
\hline
Obj&Shape&$<$10
&\cellcolor{blue!72}{\textcolor{white}{0.114}}&150&0
&\cellcolor{blue!80}{\textcolor{white}{0.032}}\\
Obj&Shape&10--50
&\cellcolor{blue!33}{\textcolor{black}{0.411}}&568&32
&\cellcolor{blue!50}{\textcolor{black}{0.164}}\\
\hline
Obj&Txtr &$<$10
&\cellcolor{blue!57}{\textcolor{white}{0.193}}&475&5
&\cellcolor{blue!61}{\textcolor{white}{0.073}}\\
Obj&Txtr &10--50
&\cellcolor{blue!20}{\textcolor{black}{0.474}}&1821&99
&\cellcolor{blue!48}{\textcolor{black}{0.158}}\\
\bottomrule
\end{tabular}

\vspace{-10pt} 
\end{table}

\subsection{Attention over Vision Tokens}

\begin{figure*}[h]
    \centering
    \includegraphics[width=0.75\textwidth]{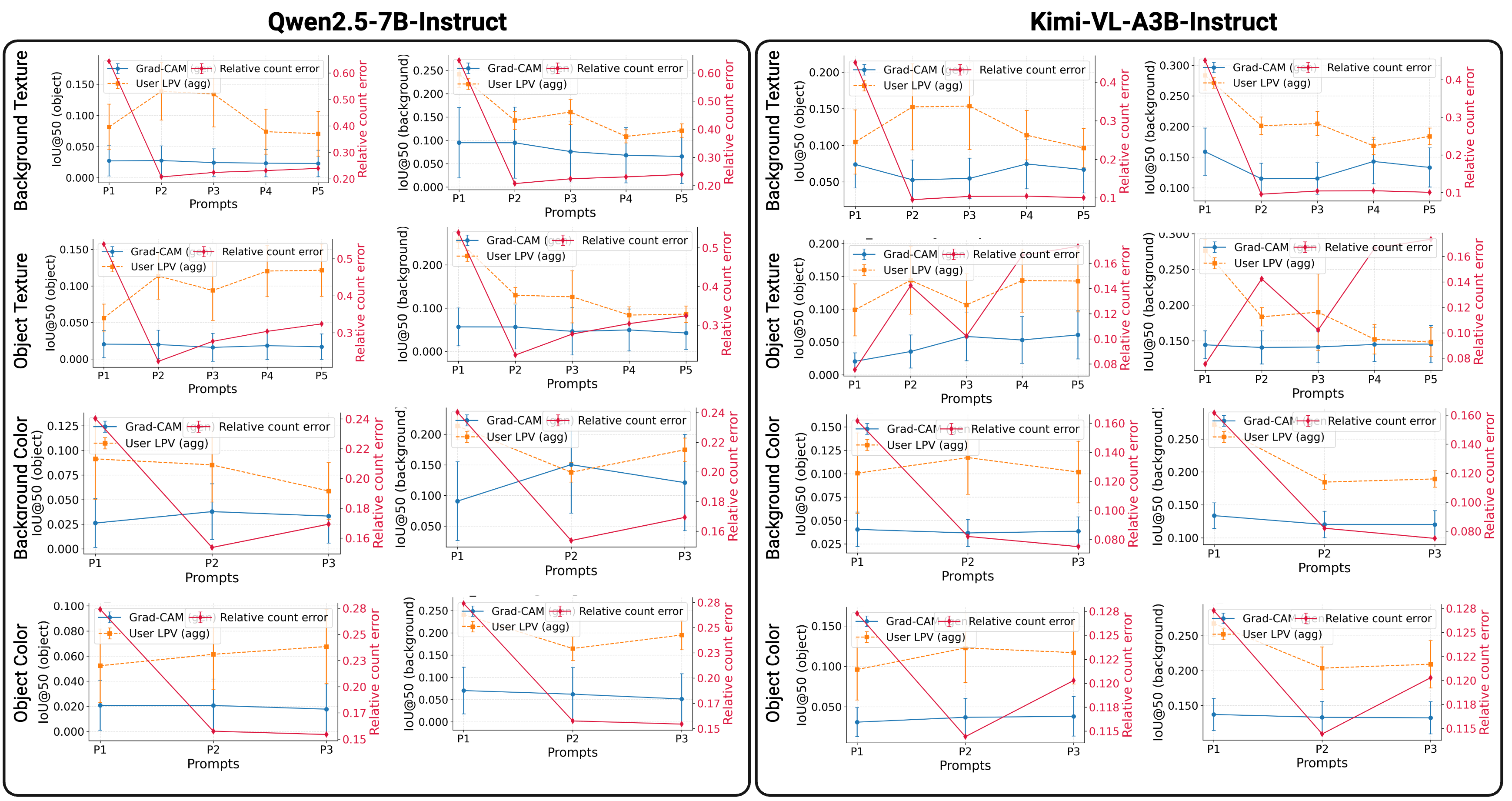}
    \caption{Visualization of the model's attention over object and background for all image types and increasing linguistic specificity in the prompt.}
    \label{fig:pattern_visual_attention}
\end{figure*}
\begin{figure*}[h]
    \centering
    \includegraphics[width=0.75\textwidth]{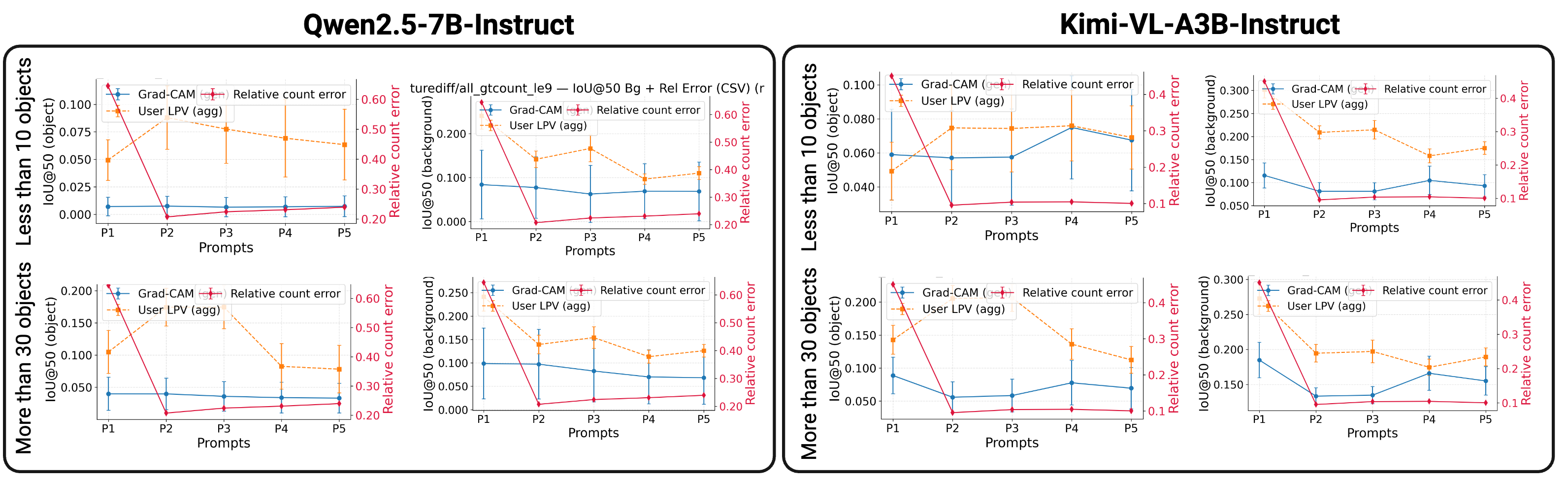}
    \caption{Visualization of the model’s attention across different background-texture patterns for images containing fewer than 10 objects or more than 30 objects.}
    \label{fig:pattern_visual_attention_count}
\end{figure*}
To investigate how linguistic variations influence the model’s spatial focus, we analyze the distribution of attention over vision tokens using five prompts for texture variations and three prompts for color variations applied to both the background and the object where each successive prompt adds more linguistic detail following our ``prompt ladder" paradigm.
In Figure \ref{fig:pattern_visual_attention}, we present the \textit{IoU@50} overlap between the attention heatmap and image for both object and background. Both Qwen2.5-7B-Instruct and Kimi-VL-A3B-Instruct exhibit a consistent trend. For Background Texture images, prompts with moderate object-related detail (P2–P3) increase LPV overlap with object regions \textit{(IoU@50, object)}, while decreasing LPV overlap with background regions \textit{(IoU@50, background)}, relative to the baseline prompt (P1). These prompts (P2–P3) introduce explicit object descriptors such as color and shape, which help the models better localize the target regions—consistent with the performance gains under moderate specificity in Section 4.1.  In contrast, prompts including additional background-related information (P4–P5) generally reduce object Grad-CAM attention \textit{(IoU@50, object)} compared to the base prompt, particularly in the Background Texture condition for both models.
For Object Texture, LPV overlap with the object consistently shows enhanced localization across P2–P5, reflecting stronger grounding when object-specific cues are present. However, beyond P3, the inclusion of fine-grained texture descriptors does not further reduce the relative count error; instead, error tends to rise again—aligning with the ``cognitive sink" effect from Section 4.1. These results suggest that linguistic specificity aids in reducing counting error up to a moderate level, but excessive descriptive detail can counteract this benefit, highlighting a non-linear relationship between prompt richness and counting error reduction.
Figure \ref{fig:pattern_visual_attention_count} shows that the number of objects in the image does not substantially influence the overall trend of visual attention. Both models exhibit consistent attention patterns for images containing fewer than 10 objects and for those with more than 30. In both settings, Grad-CAM and LPV attention distributions follow similar trajectories across prompts, though relative count error remains significantly lower in the $<$10 object condition compared to the $>$30 objects setting. This stability indicates that the models’ spatial grounding behavior, as guided by linguistic cues, is largely invariant to object density. 
The model trends are presented in the supplementary material.

\vspace{-0.25cm}
\subsection{Impact of Attention Redistribution}

\begin{table*}[h]
\centering
\caption{Attention reweighting strategy performance on synthetic (background texture) and real-world (FSC-147) datasets for Qwen3-VL-8B (dense) and KimiVL-A3B (MoE). Results are mean MRCE and Accuracy across count buckets. \textbf{Bold} values indicate improvement over baseline. Arrows indicate direction of improvement ($\downarrow$ lower is better, $\uparrow$ higher is better). ``-'' denotes undefined due to model collapse.}
\label{tab:attention_strategies}
\resizebox{\textwidth}{!}{%
\begin{tabular}{llcccccccc}
\toprule
\textbf{Group} & \textbf{Strategy} 
  & \multicolumn{2}{c}{\textbf{Qwen (Synthetic)}} 
  & \multicolumn{2}{c}{\textbf{Kimi (Synthetic)}} 
  & \multicolumn{2}{c}{\textbf{Qwen (FSC-147)}} 
  & \multicolumn{2}{c}{\textbf{Kimi (FSC-147)}} \\
\cmidrule(lr){3-4} \cmidrule(lr){5-6} \cmidrule(lr){7-8} \cmidrule(lr){9-10}
 & & MRCE$\downarrow$ & Acc$\uparrow$ & MRCE$\downarrow$ & Acc$\uparrow$ & MRCE$\downarrow$ & Acc$\uparrow$ & MRCE$\downarrow$ & Acc$\uparrow$ \\
\midrule

\textit{Baseline} 
  & baseline 
  & 0.080 & 0.330 & 0.140 & 0.320 & 0.168 & 0.220 & 0.301 & 0.160 \\

\midrule

\multirow{4}{*}{\textit{Uniform}}
  & uniform\_suppress 
  & 0.160 & 0.100 & \textbf{0.090} & 0.200 & 0.226 & 0.120 & 0.326 & 0.140 \\
  & uniform\_amplify 
  & 0.250 & 0.070 & 0.350 & 0.160 & 0.192 & \textbf{0.234} & 0.347 & \textbf{0.200} \\
  & uniform\_balance 
  & 0.080 & \textbf{0.340} & 0.320 & 0.210 & \textbf{0.158} & 0.163 & 0.694 & \textbf{0.220} \\
  & uniform\_focus 
  & ---   & 0.000 & 0.860 & 0.020 & ---   & 0.000 & 0.374 & \textbf{0.200} \\

\midrule

\textit{Alternating}
  & alternating\_amp\_sup 
  & 0.090 & 0.290 & \textbf{0.090} & 0.240 & 0.176 & 0.220 & 0.332 & 0.160 \\

\midrule

\multirow{2}{*}{\textit{Progressive}}
  & progressive\_visual\_grow 
  & 0.080 & \textbf{0.370} & 0.160 & 0.260 & \textbf{0.163} & \textbf{0.240} & 0.694 & \textbf{0.260} \\
  & progressive\_visual\_fade 
  & 0.080 & \textbf{0.370} & 0.320 & 0.200 & 0.218 & 0.163 & 0.738 & \textbf{0.200} \\

\midrule

\multirow{2}{*}{\textit{Early (image-naive)}}
  & early\_visual\_only 
  & 0.610 & 0.040 & \textbf{0.100} & 0.210 & ---   & 0.000 & 0.429 & \textbf{0.180} \\
  & extreme\_visual\_early 
  & ---   & 0.000 & 0.690 & 0.040 & ---   & 0.000 & 0.771 & 0.160 \\

\midrule

\multirow{2}{*}{\textit{Early (mask-guided)}}
  & early\_amplify\_visual\_mask 
  & 0.080 & \textbf{0.350} & 0.150 & 0.290 & \textbf{0.143} & \textbf{0.224} & 0.344 & \textbf{0.184} \\
  & early\_amplify\_visual\_mask\_bg\_suppress 
  & 0.080 & \textbf{0.360} & 0.150 & 0.290 & \textbf{0.160} & 0.200 & 0.344 & \textbf{0.184} \\

\midrule

\textit{Middle (image-naive)}
  & middle\_visual\_boost 
  & 0.230 & 0.080 & 0.420 & 0.170 & 0.185 & 0.208 & 0.307 & \textbf{0.280} \\

\midrule

\multirow{2}{*}{\textit{Middle (mask-guided)}}
  & middle\_amplify\_visual\_mask 
  & 0.270 & 0.110 & 0.220 & 0.240 & 0.195 & \textbf{0.224} & \textbf{0.281} & \textbf{0.163} \\
  & middle\_amplify\_visual\_mask\_bg\_suppress 
  & 0.300 & 0.100 & 0.220 & 0.240 & 0.225 & 0.120 & \textbf{0.281} & \textbf{0.163} \\

\midrule

\multirow{2}{*}{\textit{Late (image-naive)}}
  & late\_visual\_retention 
  & 0.090 & 0.330 & 0.280 & 0.200 & \textbf{0.145} & \textbf{0.224} & 0.725 & \textbf{0.240} \\
  & extreme\_text\_late 
  & 0.080 & \textbf{0.340} & 0.210 & 0.240 & \textbf{0.164} & 0.160 & \textbf{0.297} & \textbf{0.220} \\

\midrule

\multirow{2}{*}{\textit{Late (mask-guided)}}
  & late\_amplify\_visual\_mask 
  & 0.080 & 0.330 & 0.160 & 0.300 & \textbf{0.163} & 0.204 & \textbf{0.296} & 0.143 \\
  & \textbf{late\_amplify\_visual\_mask\_bg\_suppress} 
  & \textbf{0.070} & \textbf{0.350} & 0.160 & 0.300 & 0.172 & \textbf{0.260} & \textbf{0.296} & 0.143 \\

\bottomrule
\end{tabular}%
}
\end{table*}

We present the results of our layer wise attention reweighting experiments in Table \ref{tab:attention_strategies}. These interventions were tested on the  \texttt{background texture} dataset and, to assess the transferability of our findings to natural images, on the FSC-147 counting benchmark, with results averaged over all images.  We modify self-attention values either uniformly across all layers or in layer groups:  \texttt{early} (layers 0-7 for Qwen, 0-8 for Kimi),  \texttt{middle} (layers 8-24 for Qwen, 9-17 for Kimi),  and \texttt{late} (layers 24-31 for Qwen, 18-26 for Kimi). Refer to Supplementary Sections \ref{sec:attn_reweighting_supp} and \ref{sec:attn_reweighting_exp_supp} for detailed explanations of the experiments.

Across both synthetic and FSC-147 evaluations, mask-guided late-layer amplification emerges as the most consistent intervention: \texttt{late amplify visual mask bg suppress} is the only strategy to improve Qwen synthetic MRCE while also boosting FSC-147 accuracy, and \texttt{early amplify visual mask} similarly transfers to real-world data by improving both Qwen FSC-147 MRCE and accuracy. In contrast, image-naive strategies are largely model-specific — \texttt{uniform suppress} and \texttt{alternating amp sup} substantially reduce Kimi synthetic MRCE (0.09 vs. 0.14 baseline) but degrade Qwen, suggesting Kimi's MoE architecture is more tolerant of global suppression than Qwen's dense design. Middle-layer interventions reveal an MRCE-accuracy tradeoff, improving Kimi FSC-147 MRCE while sharply degrading Qwen accuracy. Progressive and uniform strategies produce inconsistent cross-model effects, with \texttt{uniform balance} improving Qwen FSC-147 MRCE while catastrophically inflating Kimi's. Most strikingly, extreme attention redistribution strategies cause complete model collapse on Qwen while leaving Kimi comparatively unaffected. Taken together, these results indicate that spatial object mask guidance is the critical factor for robust, generalizable attention intervention — image-naive strategies may shift error distributions without reliably improving counting.
\section{Conclusions and Future Work}
\label{sec:conclusion}

We presented a systematic diagnostic study of VLM counting capabilities, introducing a controlled synthetic benchmark to isolate the effects of prompt specificity, visual properties, object count range, and attention distribution. Our prompt ladder experiments reveal that counting failures are not primarily driven by compositional reasoning, but by the cognitive load imposed by the prompt's primary segmentation cue. Attention analysis corroborates this, showing that models can perceptually attend to objects yet fail to enumerate them when competing semantic cues overwhelm the counting process. Our intervention experiments further show that mask-guided attention amplification in early and late decoder layers is the most robust strategy for improving counting accuracy, generalizing from synthetic stimuli to real-world FSC-147 images, while image-naive strategies remain model-specific and unreliable.
Our findings motivate developing specialized attention mechanisms tailored to decoder architectures, and addressing the dissociation between counting accuracy and MRCE through hybrid training objectives. Interpretable probes could further identify network components responsible for enumeration versus classification, offering a path toward architectures whose functional organization more closely reflects the modular structure of human visual cognition, providing crucial insights for architectural improvements. 

\textbf{Limitations:} While our controlled synthetic benchmarks enable precise failure mode analysis, extending this framework to real-world scenarios with occlusion, varying scales, and complex spatial arrangements remains essential. 
Our attention interventions assume standard transformer architectures; if frontier models employ attention variants (e.g., mixture of depths or sparse attention patterns), our strategies may not directly transfer.
Finally, our analysis is limited to open-source VLMs; proprietary and larger-scale models may exhibit qualitatively different failure modes and responses to attention intervention.
{
    \small
    \bibliographystyle{ieeenat_fullname}
    \bibliography{main}
}

\clearpage
\setcounter{page}{1}
\maketitlesupplementary


\section{Layer-wise Propagation of Visual attention}
\label{sec:lpv}
\paragraph{Gradient-weighted attention.}
Inspired by \citet{Chefer_2021_ICCV} propose, we propose a lightweight gradient-weighted relevance propagation(LPV) for autoregressive VLMs that turns layer-wise attentions into token-level relevance maps by using gradient weighting and cross-layer diffusion. 
For each Transformer layer $\ell$, let $A^{(\ell)} \in \mathbb{R}^{H \times S \times S}$ be the multi-head attention (post-softmax) and let
\[
G^{(\ell)} \;=\; \frac{\partial \mathcal{L}}{\partial A^{(\ell)}}
\]
be its gradient obtained from a token-level cross-entropy loss on selected output positions. 
Noisy negative signals in the gradient are suppressed using a ReLU: 
\[
\tilde{G}^{(\ell)} \;=\; \mathrm{ReLU}\!\left(G^{(\ell)}\right).
\]
We then form a gradient-weighted attention map by element-wise interaction and heads averaging:
\[
H^{(\ell)} \;=\;
\frac{1}{H} \sum_{h=1}^{H}
\bigl(A^{(\ell)}_{h} \odot \tilde{G}^{(\ell)}_{h}\bigr).
\]
To preserve self-information paths, we add the identity and apply row-normalize to obtain a row-stochastic per-layer relevance transition:
\[
M^{(\ell)}(i,j) \;=\;
\frac{H^{(\ell)}(i,j) + \delta_{ij}}
     {\sum_{k}\bigl(H^{(\ell)}(i,k)+\delta_{ik}\bigr)}.
\]

\paragraph{Cross-layer joint relevance.}
We compose the last $K$ layers to aggregate both deep semantics and shallow localization:
\[
C \;=\; \prod_{\ell=L-K+1}^{L} M^{(\ell)}.
\]
For any given output token at index $t$, the corresponding row  $C[t,:]$ gives a fine-grained relevance distribution over all input tokens (textual and visual) that contribute to the prediction of that specific token.  

\paragraph{Differences from \citet{Chefer_2021_ICCV}.}
Our formulation follows the gradient-weighted attention idea and cross-layer composition of Chefer et al., but is tailored to autoregressive VLMs:
(1) \textbf{Token-specific supervision.} We supervise a token-level cross-entropy on a \emph{selected} set of decoding steps and collect $\{\nabla A^{(\ell)}\}$ in a single backward pass, enabling token/time-step-specific attribution and natural multi-token aggregation.
(2) \textbf{Per-layer row-stochastic transitions.} We explicitly enforce row-stochastic $M^{(\ell)}$ by adding the identity and applying row normalization at each layer, which stabilizes deep products and improves robustness.
(3) \textbf{Controllable depth.} We optionally compose only the last $K$ layers to trade interpretability depth for stability and speed without modifying model internals.

\paragraph{Why not plain attention rollout?}
Our method provides more faithful and selective relevance maps compared to simpler methods like Attention Rollout. Plain Attention rollout composes raw attentions and is class-/token-agnostic. It highlights tokens the model "looked at" (high $A$), but not necessarily tokens that \textit{influenced} the specific prediction. 
Our gradient-weighted diffusion emphasizes attention edges   that are both \emph{used} (large $A$) and \emph{useful for the current prediction} (large positive $\nabla A$), yielding more selective and faithful relevance maps that adapt naturally to multi-token, multi-modal settings.

\section{Attention Reweighting in Qwen Models}

\subsection{Attention Reweighting in Grouped Query Attention Architecture}
\label{sec:attn_reweighting_supp}

Qwen 2.5 and Qwen 3 models employ Grouped Query Attention (GQA) 
, which differs from standard Multi-Head Attention by using fewer key-value heads than query heads to reduce computational cost. Specifically, with $H=32$ attention heads and $K=8$ key-value heads, each KV head is shared across $G = H/K = 4$ query heads. This architecture necessitates special handling during attention reweighting. After projecting inputs through $W_q$, $W_k$, and $W_v$, the resulting tensors have shapes $\mathbf{Q} \in \mathbb{R}^{B \times L \times Hd}$ and $\mathbf{K}, \mathbf{V} \in \mathbb{R}^{B \times L \times Kd}$, where $B$ is batch size, $L$ is sequence length, and $d$ is head dimension. Before computing attention, we reshape these to $\mathbf{Q} \in \mathbb{R}^{B \times H \times L \times d}$ and $\mathbf{K}, \mathbf{V} \in \mathbb{R}^{B \times K \times L \times d}$. The key-value heads must then be repeated via $\text{repeat\_kv}(\mathbf{V}, G)$ to obtain $\mathbf{V}' \in \mathbb{R}^{B \times H \times L \times d}$ by expanding along a new dimension and reshaping: $\mathbf{V}' = \text{reshape}(\mathbf{V}[:,:,\text{None},:,:].expand(B, K, G, L, d), [B, H, L, d])$. This ensures dimensional compatibility when applying reweighted attention weights $\tilde{\mathbf{A}} \in \mathbb{R}^{B \times H \times L \times L}$ to compute the output $\mathbf{O} = \tilde{\mathbf{A}} \mathbf{V}'$. Our implementation maintains a full cache of value projections across generation steps, repeats the KV heads appropriately, applies the reweighting strategy to the attention weights, recomputes the attention output with modified weights, and finally applies the output projection $W_o$. This approach preserves the efficiency benefits of GQA while enabling fine-grained control over visual-textual attention distribution across all $H$ query heads.

\subsection{Attention Reweighting Experiments}
\label{sec:attn_reweighting_exp_supp}

\begin{table*}[]
\centering
\caption{Qwen 3-VL-8B-Instruct MRCE and Accuracy results for different object count buckets averaged over the \texttt{background texture} dataset. Results sorted by Mean MRCE over all buckets. We see that object mask guided attention reweighting methods beat baseline.}
\label{tab:acc_mrce_by_count_q3}
\resizebox{\textwidth}{!}{%
\begin{tabular}{lllllllllllll}
\hline
\multicolumn{1}{c}{\textbf{Attn. Strat.}}   & \multicolumn{2}{c}{\textbf{0-10}} & \multicolumn{2}{c}{\textbf{10-20}} & \multicolumn{2}{c}{\textbf{20-30}} & \multicolumn{2}{c}{\textbf{30-40}} & \multicolumn{2}{c}{\textbf{40-50}} & \multicolumn{2}{c}{\textbf{Mean}} \\
                                            & \textbf{MRCE}   & \textbf{Acc.}   & \textbf{MRCE}    & \textbf{Acc.}   & \textbf{MRCE}    & \textbf{Acc.}   & \textbf{MRCE}    & \textbf{Acc.}   & \textbf{MRCE}    & \textbf{Acc.}   & \textbf{MRCE}   & \textbf{Acc.}   \\ \hline
late\_amplify\_visual\_mask\_bg\_suppress   & 0.02            & 0.87            & 0.05             & 0.5             & 0.09             & 0.17            & 0.09             & 0.11            & 0.12             & 0.11            & 0.07            & 0.35            \\
late\_amplify\_visual\_mask                 & 0.02            & 0.87            & 0.05             & 0.44            & 0.1              & 0.13            & 0.09             & 0.15            & 0.14             & 0.06            & 0.08            & 0.33            \\
baseline                                    & 0.02            & 0.87            & 0.05             & 0.42            & 0.09             & 0.2             & 0.1              & 0.12            & 0.14             & 0.06            & 0.08            & 0.33            \\
early\_amplify\_visual\_mask                & 0.02            & 0.89            & 0.05             & 0.47            & 0.09             & 0.2             & 0.1              & 0.13            & 0.14             & 0.06            & 0.08            & 0.35            \\
early\_amplify\_visual\_mask\_bg\_suppress  & 0.01            & 0.91            & 0.05             & 0.45            & 0.08             & 0.25            & 0.1              & 0.1             & 0.15             & 0.08            & 0.08            & 0.36            \\
extreme\_text\_late                         & 0.02            & 0.85            & 0.05             & 0.43            & 0.1              & 0.18            & 0.08             & 0.17            & 0.15             & 0.08            & 0.08            & 0.34            \\
uniform\_balance                            & 0.02            & 0.88            & 0.05             & 0.43            & 0.09             & 0.14            & 0.1              & 0.14            & 0.15             & 0.11            & 0.08            & 0.34            \\
progressive\_visual\_fade                   & 0.02            & 0.85            & 0.04             & 0.48            & 0.08             & 0.25            & 0.09             & 0.13            & 0.18             & 0.16            & 0.08            & 0.37            \\
progressive\_visual\_grow                   & 0.02            & 0.9             & 0.04             & 0.49            & 0.09             & 0.22            & 0.1              & 0.16            & 0.17             & 0.06            & 0.08            & 0.37            \\
late\_visual\_retention                     & 0.02            & 0.86            & 0.05             & 0.41            & 0.08             & 0.19            & 0.11             & 0.12            & 0.18             & 0.09            & 0.09            & 0.33            \\
alternating\_amp\_sup                       & 0.03            & 0.77            & 0.07             & 0.28            & 0.1              & 0.16            & 0.09             & 0.15            & 0.17             & 0.07            & 0.09            & 0.29            \\
uniform\_suppress                           & 0.1             & 0.42            & 0.17             & 0               & 0.2              & 0               & 0.17             & 0.02            & 0.14             & 0.06            & 0.16            & 0.1             \\
middle\_visual\_boost                       & 0.13            & 0.27            & 0.18             & 0.05            & 0.2              & 0.04            & 0.28             & 0.01            & 0.35             & 0.05            & 0.23            & 0.08            \\
uniform\_amplify                            & 0.14            & 0.25            & 0.2              & 0.05            & 0.22             & 0.01            & 0.29             & 0               & 0.38             & 0.03            & 0.25            & 0.07            \\
middle\_amplify\_visual\_mask               & 0.1             & 0.39            & 0.17             & 0.09            & 0.22             & 0.06            & 0.32             & 0.01            & 0.54             & 0.01            & 0.27            & 0.11            \\
middle\_amplify\_visual\_mask\_bg\_suppress & 0.1             & 0.38            & 0.17             & 0.1             & 0.26             & 0.01            & 0.36             & 0.01            & 0.6              & 0.01            & 0.3             & 0.1             \\
early\_visual\_only                         & 0.7             & 0               & 0.26             & 0.19            & 0.62             & 0               & 0.72             & 0               & 0.78             & 0               & 0.61            & 0.04            \\
extreme\_visual\_early                      & NaN             & 0               & NaN              & 0.08            & NaN              & 0               & NaN              & 0               & NaN              & 0               & NaN             & 0.02            \\
uniform\_focus                              & NaN             & 0               & NaN              & 0               & NaN              & 0               & NaN              & 0               & NaN              & 0               & NaN             & 0               \\ \hline
\end{tabular}%
}
\end{table*}

\begin{table*}[]
\centering
\caption{Qwen 2.5 VL-7B-Instruct MRCE and Accuracy results for different object count buckets averaged over the \texttt{background texture} dataset. Results sorted by Mean MRCE over all buckets. We see that object mask guided attention reweighting methods and methods like \texttt{alternating\_amp\_sup} beat \texttt{baseline} in MRCE. As object counts increase, attention reweighting methods lead to modest gains from baseline performance.}
\label{tab:acc_mrce_by_count_q2_5}
\resizebox{\textwidth}{!}{%
\begin{tabular}{lllllllllllll}
\hline
\multicolumn{1}{c}{\textbf{Attn. Strat.}}   & \multicolumn{2}{c}{\textbf{0-10}} & \multicolumn{2}{c}{\textbf{10-20}} & \multicolumn{2}{c}{\textbf{20-30}} & \multicolumn{2}{c}{\textbf{30-40}} & \multicolumn{2}{c}{\textbf{40-50}} & \multicolumn{2}{c}{\textbf{Mean}} \\
                                            & \textbf{MRCE}   & \textbf{Acc.}   & \textbf{MRCE}    & \textbf{Acc.}   & \textbf{MRCE}    & \textbf{Acc.}   & \textbf{MRCE}    & \textbf{Acc.}   & \textbf{MRCE}    & \textbf{Acc.}   & \textbf{MRCE}   & \textbf{Acc.}   \\ \hline
early\_amplify\_visual\_mask                & 0.08            & 0.61            & 0.13             & 0.12            & 0.11             & 0.12            & 0.15             & 0.12            & 0.21             & 0.03            & 0.14            & 0.2             \\
early\_amplify\_visual\_mask\_bg\_suppress  & 0.09            & 0.57            & 0.13             & 0.11            & 0.1              & 0.15            & 0.15             & 0.12            & 0.22             & 0.01            & 0.14            & 0.19            \\
late\_amplify\_visual\_mask\_bg\_suppress   & 0.08            & 0.61            & 0.12             & 0.15            & 0.1              & 0.12            & 0.17             & 0.08            & 0.22             & 0.01            & 0.14            & 0.19            \\
alternating\_amp\_sup                       & 0.09            & 0.57            & 0.12             & 0.15            & 0.1              & 0.13            & 0.17             & 0.08            & 0.22             & 0.02            & 0.14            & 0.19            \\
progressive\_visual\_grow                   & 0.09            & 0.59            & 0.12             & 0.08            & 0.1              & 0.15            & 0.16             & 0.11            & 0.23             & 0.01            & 0.14            & 0.19            \\
late\_amplify\_visual\_mask                 & 0.08            & 0.61            & 0.12             & 0.15            & 0.1              & 0.12            & 0.17             & 0.09            & 0.22             & 0               & 0.14            & 0.19            \\
middle\_amplify\_visual\_mask               & 0.09            & 0.59            & 0.14             & 0.11            & 0.11             & 0.09            & 0.15             & 0.11            & 0.2              & 0.05            & 0.14            & 0.19            \\
baseline                                    & 0.08            & 0.61            & 0.12             & 0.14            & 0.1              & 0.12            & 0.17             & 0.09            & 0.22             & 0               & 0.14            & 0.19            \\
middle\_amplify\_visual\_mask\_bg\_suppress & 0.09            & 0.57            & 0.14             & 0.05            & 0.11             & 0.13            & 0.16             & 0.09            & 0.2              & 0.05            & 0.14            & 0.18            \\
progressive\_visual\_fade                   & 0.13            & 0.58            & 0.15             & 0.1             & 0.11             & 0.1             & 0.14             & 0.15            & 0.19             & 0.06            & 0.14            & 0.2             \\
uniform\_balance                            & 0.12            & 0.57            & 0.13             & 0.14            & 0.1              & 0.15            & 0.16             & 0.09            & 0.21             & 0.04            & 0.14            & 0.2             \\
late\_visual\_retention                     & 0.12            & 0.57            & 0.13             & 0.14            & 0.1              & 0.14            & 0.16             & 0.09            & 0.21             & 0.04            & 0.14            & 0.19            \\
extreme\_text\_late                         & 0.13            & 0.54            & 0.13             & 0.16            & 0.1              & 0.14            & 0.16             & 0.08            & 0.22             & 0.02            & 0.15            & 0.19            \\
middle\_visual\_boost                       & 0.11            & 0.53            & 0.21             & 0.05            & 0.15             & 0.06            & 0.12             & 0.2             & 0.16             & 0.11            & 0.15            & 0.19            \\
uniform\_amplify                            & 0.12            & 0.52            & 0.23             & 0.07            & 0.19             & 0.05            & 0.11             & 0.22            & 0.15             & 0.1             & 0.16            & 0.19            \\
uniform\_suppress                           & 0.1             & 0.54            & 0.17             & 0.02            & 0.14             & 0.09            & 0.22             & 0.03            & 0.27             & 0.01            & 0.18            & 0.14            \\
early\_visual\_only                         & NaN             & 0               & NaN              & 0               & NaN              & 0               & NaN              & 0               & NaN              & 0               & NaN             & 0               \\
extreme\_visual\_early                      & NaN             & 0               & NaN              & 0               & NaN              & 0               & NaN              & 0               & NaN              & 0               & NaN             & 0               \\
uniform\_focus                              & NaN             & 0               & NaN              & 0               & NaN              & 0               & NaN              & 0               & NaN              & 0               & NaN             & 0               \\ \hline
\end{tabular}%
}
\end{table*}

We experiment with more comprehensive layer based configurations to measure the impact of changing attention values over different layers. In all experiments we use the same prompt: \texttt{"Count the number of objects in this image. Answer the count within curly brackets, eg. \{10\}"}. We used only the \texttt{background texture} dataset for this evaluation because that provided a more challenging visual use case.

\subsubsection{Experimental Configurations}

\paragraph{Layer Groups.}
\begin{itemize}
    \item \textbf{Early (0--7)}: Feature extraction---low-level visual patterns and syntactic structure
    \item \textbf{Middle (8--23)}: Semantic integration---multimodal fusion and object relationships
    \item \textbf{Late (24--31)}: High-level reasoning---global reasoning and linguistic refinement
\end{itemize}

We evaluate 19 attention reweighting configurations across six categories:

\paragraph{Baseline.}
\begin{itemize}
    \item \texttt{baseline}: No modification---standard model inference
\end{itemize}

\paragraph{Uniform Strategies (all layers).}
\begin{itemize}
    \item \texttt{uniform\_amplify}: $2\times$ visual attention throughout
    \item \texttt{uniform\_suppress}: $0.5\times$ visual attention throughout
    \item \texttt{uniform\_focus}: Exclusive visual attention throughout
    \item \texttt{uniform\_balance}: Maintained 40\% visual ratio throughout
\end{itemize}

\paragraph{Progressive Strategies (layer-wise transitions).}
\begin{itemize}
    \item \texttt{progressive\_visual\_fade}: Strong $\rightarrow$ balanced $\rightarrow$ weak visual (amplify/balance/suppress)
    \item \texttt{progressive\_visual\_grow}: Weak $\rightarrow$ balanced $\rightarrow$ strong visual (suppress/balance/amplify)
\end{itemize}

\paragraph{Localized Strategies (group-specific).}
\begin{itemize}
    \item \texttt{early\_visual\_only}: Focus early layers, suppress middle-late
    \item \texttt{middle\_visual\_boost}: Amplify middle layers, balance elsewhere
    \item \texttt{late\_visual\_retention}: Amplify late layers, balance elsewhere
    \item \texttt{extreme\_visual\_early}: Focus first 37.5\% of layers (0--11), balance rest
    \item \texttt{extreme\_text\_late}: Suppress final 37.5\% of layers (20--31), balance rest
\end{itemize}

\paragraph{Alternating Strategy.}
\begin{itemize}
    \item \texttt{alternating\_amp\_sup}: Layer-by-layer amplify/suppress alternation
\end{itemize}

\paragraph{Object-Aware Strategies (segmentation-guided).}
\begin{itemize}
    \item \texttt{early/middle/late\_amplify\_visual\_mask}: Object amplification ($2\times$) with no background suppression in respective layer groups
    \item \texttt{early/middle/\\
    late\_amplify\_visual\_mask\_bg\_suppress}: Object amplification ($2\times$) with strong background suppression ($0.5\times$) in respective layer groups
\end{itemize}

\begin{table*}[h]
\centering
\caption{KimiVL-A3B-Instruct MRCE and Accuracy results for different object count buckets averaged over the \texttt{background texture} dataset. Results sorted by Mean MRCE over all buckets. We see that attention reweighting methods like \texttt{uniform\_suppress} and \texttt{alternating\_amp\_sup } beat \texttt{baseline} in both MRCE, but do not match accuracy.}
\label{tab:acc_mrce_by_count_kimi}
\resizebox{\textwidth}{!}{%
\begin{tabular}{lllllllllllll}
\hline
\multicolumn{1}{c}{\textbf{Attn. Strat.}} &
\multicolumn{2}{c}{\textbf{0-10}} &
\multicolumn{2}{c}{\textbf{10-20}} &
\multicolumn{2}{c}{\textbf{20-30}} &
\multicolumn{2}{c}{\textbf{30-40}} &
\multicolumn{2}{c}{\textbf{40-50}} &
\multicolumn{2}{c}{\textbf{Mean}} \\
& \textbf{MRCE} & \textbf{Acc.} &
  \textbf{MRCE} & \textbf{Acc.} &
  \textbf{MRCE} & \textbf{Acc.} &
  \textbf{MRCE} & \textbf{Acc.} &
  \textbf{MRCE} & \textbf{Acc.} &
  \textbf{MRCE} & \textbf{Acc.} \\
\hline

alternating\_amp\_sup                     & 0.05 & 0.58 & 0.10 & 0.14 & 0.09 & 0.15 & 0.07 & 0.10 & 0.14 & 0.10 & 0.09 & 0.24 \\
uniform\_suppress                          & 0.07 & 0.46 & 0.10 & 0.14 & 0.06 & 0.20 & 0.11 & 0.00 & 0.10 & 0.10 & 0.09 & 0.20 \\
early\_visual\_only                        & 0.07 & 0.54 & 0.10 & 0.07 & 0.09 & 0.10 & 0.15 & 0.10 & 0.11 & 0.10 & 0.10 & 0.21 \\
baseline                                   & 0.04 & 0.73 & 0.07 & 0.43 & 0.09 & 0.15 & 0.27 & 0.10 & 0.25 & 0.10 & 0.14 & 0.32 \\
early\_amplify\_visual\_mask               & 0.05 & 0.69 & 0.07 & 0.43 & 0.12 & 0.10 & 0.32 & 0.05 & 0.20 & 0.10 & 0.15 & 0.29 \\
early\_amplify\_visual\_mask\_bg\_suppress & 0.05 & 0.69 & 0.07 & 0.43 & 0.12 & 0.10 & 0.32 & 0.05 & 0.20 & 0.10 & 0.15 & 0.29 \\
late\_amplify\_visual\_mask                & 0.04 & 0.73 & 0.08 & 0.36 & 0.09 & 0.15 & 0.32 & 0.05 & 0.30 & 0.10 & 0.16 & 0.30 \\
late\_amplify\_visual\_mask\_bg\_suppress  & 0.04 & 0.73 & 0.08 & 0.36 & 0.09 & 0.15 & 0.32 & 0.05 & 0.30 & 0.10 & 0.16 & 0.30 \\
progressive\_visual\_grow                  & 0.04 & 0.73 & 0.14 & 0.21 & 0.12 & 0.10 & 0.23 & 0.00 & 0.30 & 0.10 & 0.16 & 0.26 \\
extreme\_text\_late                        & 0.05 & 0.69 & 0.14 & 0.21 & 0.17 & 0.10 & 0.34 & 0.00 & 0.40 & 0.05 & 0.21 & 0.24 \\
middle\_amplify\_visual\_mask              & 0.05 & 0.65 & 0.16 & 0.36 & 0.20 & 0.05 & 0.39 & 0.00 & 0.35 & 0.05 & 0.22 & 0.24 \\
middle\_amplify\_visual\_mask\_bg\_suppress& 0.05 & 0.65 & 0.16 & 0.36 & 0.20 & 0.05 & 0.39 & 0.00 & 0.35 & 0.05 & 0.22 & 0.24 \\
late\_visual\_retention                    & 0.06 & 0.65 & 0.13 & 0.21 & 0.24 & 0.00 & 0.44 & 0.00 & 0.56 & 0.00 & 0.28 & 0.20 \\
uniform\_balance                           & 0.06 & 0.65 & 0.13 & 0.29 & 0.25 & 0.00 & 0.44 & 0.00 & 0.72 & 0.00 & 0.32 & 0.21 \\
progressive\_visual\_fade                  & 0.06 & 0.62 & 0.17 & 0.29 & 0.29 & 0.00 & 0.44 & 0.00 & 0.66 & 0.00 & 0.32 & 0.20 \\
uniform\_amplify                           & 0.06 & 0.62 & 0.23 & 0.00 & 0.29 & 0.00 & 0.46 & 0.00 & 0.78 & 0.00 & 0.35 & 0.16 \\
middle\_visual\_boost                      & 0.07 & 0.58 & 0.20 & 0.14 & 0.34 & 0.00 & 0.60 & 0.00 & 0.96 & 0.00 & 0.42 & 0.17 \\
extreme\_visual\_early                     & 0.17 & 0.15 & 0.36 & 0.00 & 0.80 & 0.00 & 1.16 & 0.00 & 1.03 & 0.00 & 0.69 & 0.04 \\
uniform\_focus                             & 0.29 & 0.08 & 0.64 & 0.00 & 1.24 & 0.00 & 1.10 & 0.00 & 1.13 & 0.00 & 0.86 & 0.02 \\
\hline
\end{tabular}%
}
\end{table*}

\begin{figure}[H]
    \centering

    \begin{subfigure}[t]{0.15\textwidth}
        \centering
        \includegraphics[width=\linewidth]{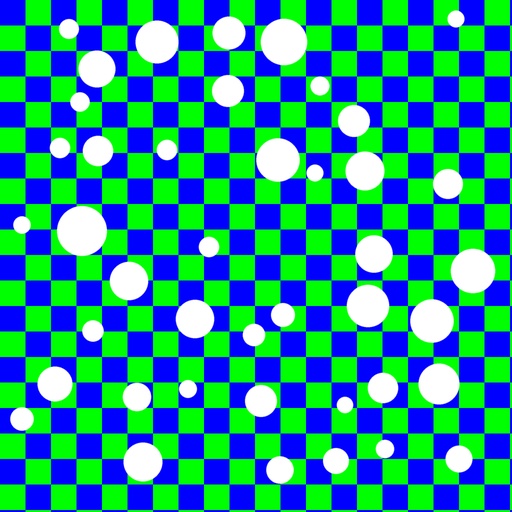}
        \caption{\protect\mbox{Checkerboard}}
    \end{subfigure}\hfill
    \begin{subfigure}[t]{0.15\textwidth}
        \centering
        \includegraphics[width=\linewidth]{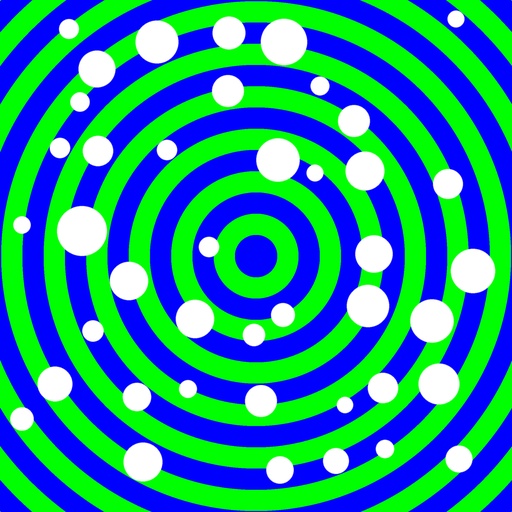}
        \caption{\protect\mbox{Concentric Rings}}
    \end{subfigure}\hfill
    \begin{subfigure}[t]{0.15\textwidth}
        \centering
        \includegraphics[width=\linewidth]{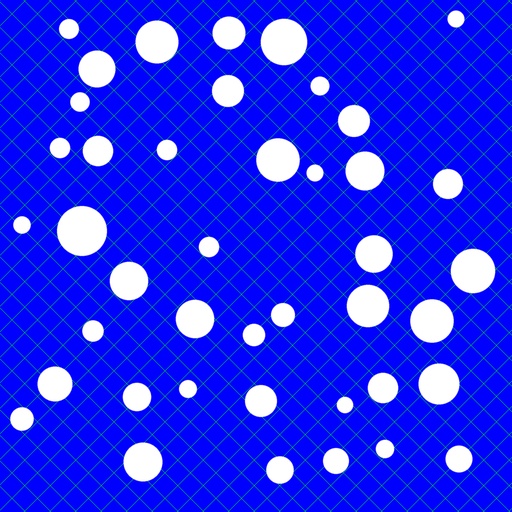}
        \caption{Crosshatch}
    \end{subfigure}

    \vspace{0.3em}
    \begin{subfigure}[t]{0.15\textwidth}
        \centering
        \includegraphics[width=\linewidth]{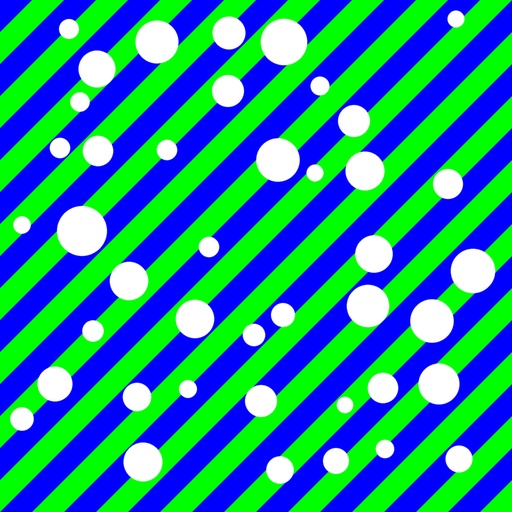}
        \caption{\protect\mbox{Diagonal Stripes}}
    \end{subfigure}\hfill
    \begin{subfigure}[t]{0.15\textwidth}
        \centering
        \includegraphics[width=\linewidth]{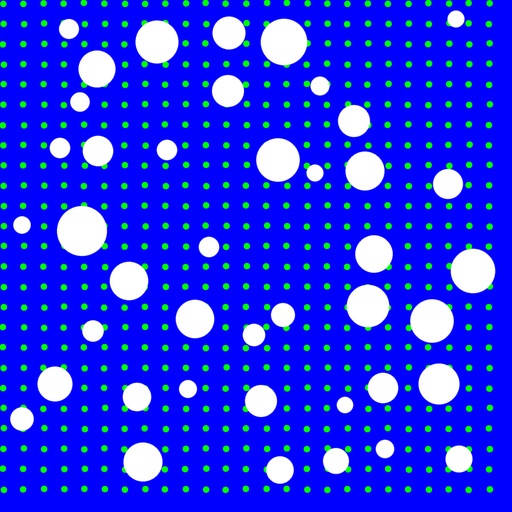}
        \caption{Dots}
    \end{subfigure}\hfill
    \begin{subfigure}[t]{0.15\textwidth}
        \centering
        \includegraphics[width=\linewidth]{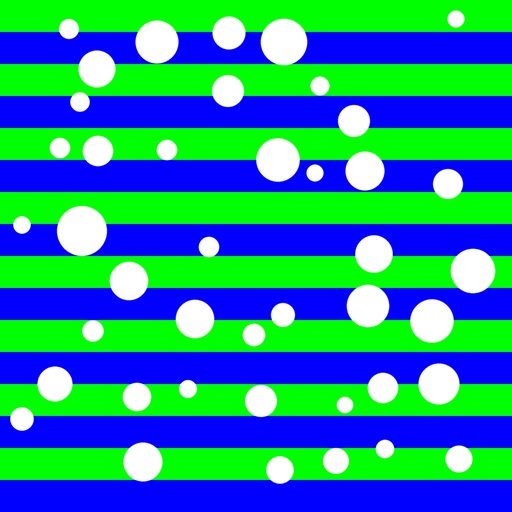}
        \caption{\protect\mbox{Horizontal Stripes}}
    \end{subfigure}

    \vspace{0.3em}
    \begin{subfigure}[t]{0.15\textwidth}
        \centering
        \includegraphics[width=\linewidth]{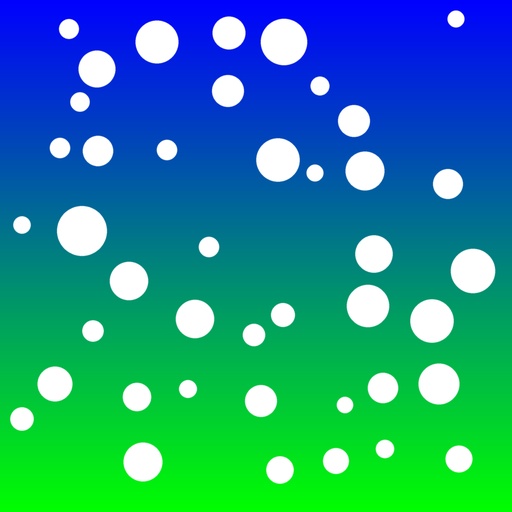}
        \caption{\protect\mbox{Linear Gradient}}
    \end{subfigure}\hfill
    \begin{subfigure}[t]{0.15\textwidth}
        \centering
        \includegraphics[width=\linewidth]{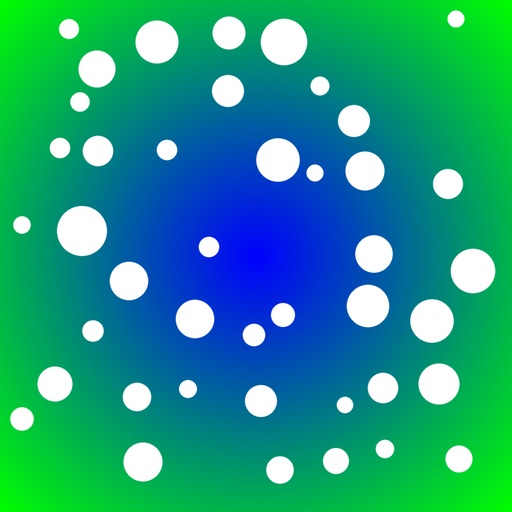}
        \caption{\protect\mbox{Radial Gradient}}
    \end{subfigure}\hfill
    \begin{subfigure}[t]{0.15\textwidth}
        \centering
        \includegraphics[width=\linewidth]{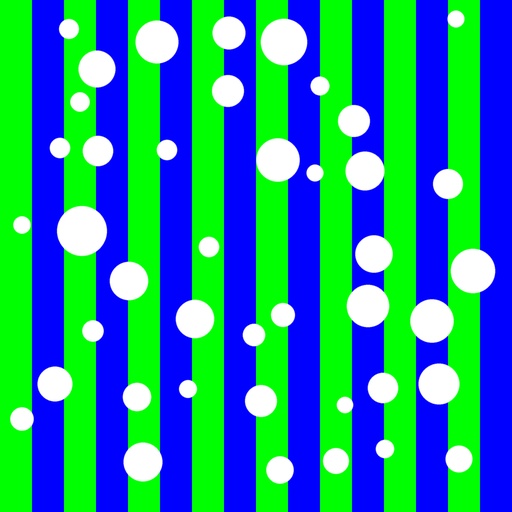}
        \caption{\protect\mbox{Vertical Stripes}}
    \end{subfigure}

    \vspace{0.3em}
    \begin{subfigure}[t]{0.15\textwidth}
        \centering
        \includegraphics[width=\linewidth]{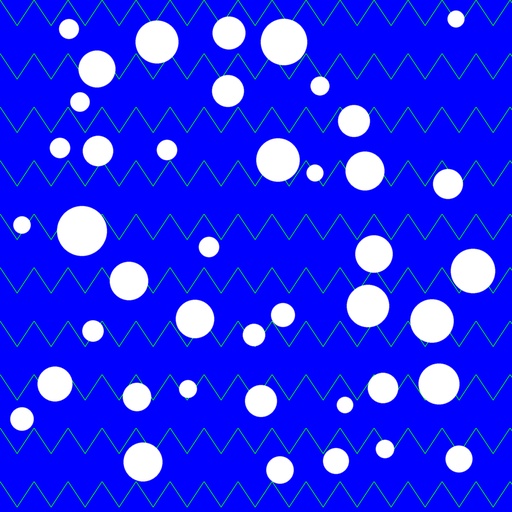}
        \caption{Zigzag}
    \end{subfigure}\hfill
    \begin{subfigure}[t]{0.15\textwidth}
        \centering
        \includegraphics[width=\linewidth]{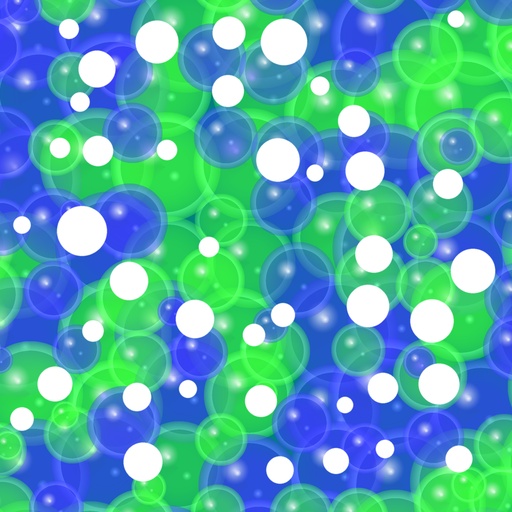}
        \caption{Bubbles}
        \label{img:bubbles}
    \end{subfigure}\hfill
    \begin{subfigure}[t]{0.15\textwidth}
        \centering
        \includegraphics[width=\linewidth]{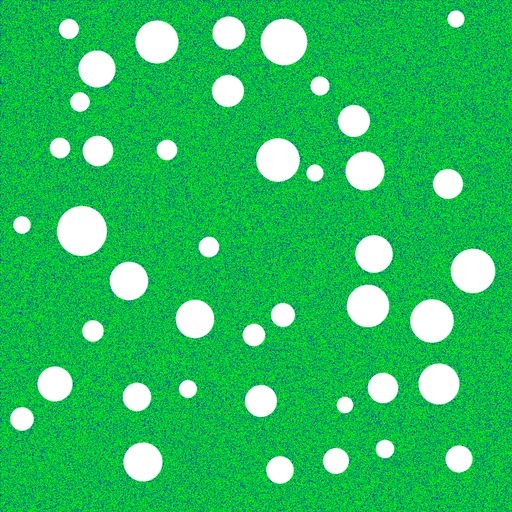}
        \caption{Noise}
    \end{subfigure}

    \caption{Example images for the \textbf{Background} category, \textbf{Texture} pattern, showing various background texture types.}
    \label{fig:bg_texture_imgs}
\end{figure}

\paragraph{Findings:}
\cref{tab:acc_mrce_by_count_q3} presents Qwen3-VL-8B-Instruct MRCE and accuracy results for the background texture dataset across different object count buckets. We do not include the \texttt{bubbles} (\cref{img:bubbles}) texture since it confuses all VLMs into counting the bubbles as objects leading to high errors.

Progressive growth strategies (\texttt{progressive\_visual\_grow}, mean MRCE 0.37) and late-stage object masked amplification with background suppression (\texttt{late\_amplify\_visual\_mask\_bg\_suppress}, mean MRCE 0.35) consistently outperform the baseline (0.33 MRCE), showing that gradual, layer-wise amplification of visual tokens while suppressing background information improves counting accuracy. The performance degradation across count buckets—from near-perfect accuracy in the 0-10 range (MRCE $\sim$0.02, Acc $\sim$0.87-0.90) to substantial errors in the 40-50 range (MRCE $\sim$0.06-0.16, Acc $\sim$0.08)—confirms that higher counts remain challenging even with attention reweighting, though our best methods maintain more graceful degradation than baseline. Critically, our experiments with aggressive early-layer interventions (\texttt{early\_visual\_only}, \texttt{extreme\_visual\_early}) demonstrate the importance of preserving text-visual alignment, as these methods catastrophically fail (mean MRCE 0.04-0.02) by over-suppressing linguistic representations. 

\cref{tab:acc_mrce_by_count_q2_5} shows results from Qwen2.5-VL-7B-Instruct. The results on the background texture dataset reveal substantially different dynamics compared to Qwen 3-VL-8B, with most attention reweighting strategies achieving marginal improvements over baseline (mean MRCE 0.14 across top performers). Object mask-guided interventions including \texttt{early\_amplify\_visual\_mask}, \texttt{late\_amplify\_visual\_mask\_bg\_suppress}, and \texttt{alternating\_amp\_sup} all converge to 0.14 mean MRCE with 0.19-0.20 accuracy, suggesting that Qwen 2.5's architecture may already incorporate more effective visual attention mechanisms that limit the potential gains from our reweighting interventions. The compressed performance distribution—with nearly all viable strategies clustering tightly around baseline—indicates that this model family exhibits greater robustness to attention modifications, though it still demonstrates the characteristic degradation pattern across count buckets (0-10 range: MRCE $\sim$0.08-0.13, Acc $\sim$0.57-0.61; 40-50 range: MRCE $\sim$0.19-0.23, Acc $\sim$0.0-0.06). Notably, the same catastrophic failure modes persist for aggressive early-layer suppression strategies (\texttt{early\_visual\_only}, \texttt{extreme\_visual\_early}, \texttt{uniform\_focus}), reinforcing our earlier finding that preserving text-visual alignment throughout the model is critical. These results suggest that while attention reweighting remains a viable intervention strategy, its effectiveness is highly architecture-dependent, with newer model families potentially incorporating design elements that already optimize attention patterns for visual reasoning tasks, thereby reducing the headroom for external manipulation.

\cref{tab:acc_mrce_by_count_kimi} shows result from Kimi-VL-A3B. The results on the background texture dataset reveal a striking divergence from the Qwen model families, with attention reweighting strategies exhibiting counterintuitive behavior where improvements in MRCE metrics fail to translate to accuracy gains and often cause catastrophic performance collapse. While \texttt{uniform\_suppress} and \texttt{alternating\_amp\_sup} achieve the lowest mean MRCE (0.09) and nominally outperform baseline (0.14 MRCE), they paradoxically underperform in accuracy (0.20-0.24 vs. 0.32 baseline), suggesting these interventions induce a counting bias that reduces error magnitude but compromises exact match performance. More concerning, many ostensibly moderate strategies—including previously successful approaches like \texttt{progressive\_visual\_grow}, \texttt{late\_amplify\_visual\_mask}, and various middle-layer interventions—cause complete accuracy collapse to 0.00 in higher count buckets (30-40 and 40-50 ranges), despite maintaining reasonable MRCE values. This pattern indicates that KimiVL's architecture is fundamentally more brittle to attention manipulation, with interventions that successfully enhance visual attention in other models disrupting critical computational pathways in this smaller (3B parameter) architecture. The extreme sensitivity extends to aggressive strategies, where \texttt{uniform\_focus} and \texttt{extreme\_visual\_early} produce catastrophic failures (0.86 and 0.69 mean MRCE respectively), reinforcing that smaller VLMs may lack the representational capacity to tolerate significant attention reweighting without losing essential cross-modal reasoning capabilities that enable accurate counting.

\section{Sample Images}
\label{sec:sample_images_supp}

\begin{figure}[!h]
    \centering
    \begin{subfigure}[b]{0.11\textwidth}
        \centering
        \includegraphics[width=\linewidth]{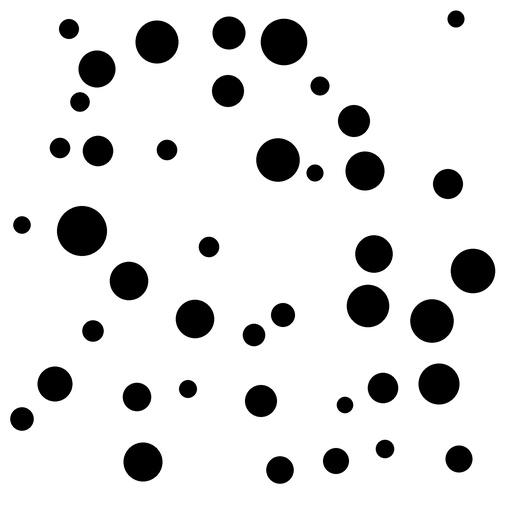}
        \caption{black}
    \end{subfigure}\hfill
    \begin{subfigure}[b]{0.11\textwidth}
        \centering
        \includegraphics[width=\linewidth]{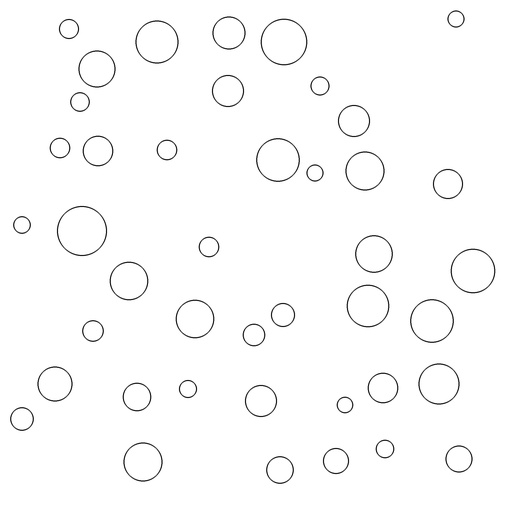}
        \caption{white}
    \end{subfigure}\hfill
    \begin{subfigure}[b]{0.11\textwidth}
        \centering
        \includegraphics[width=\linewidth]{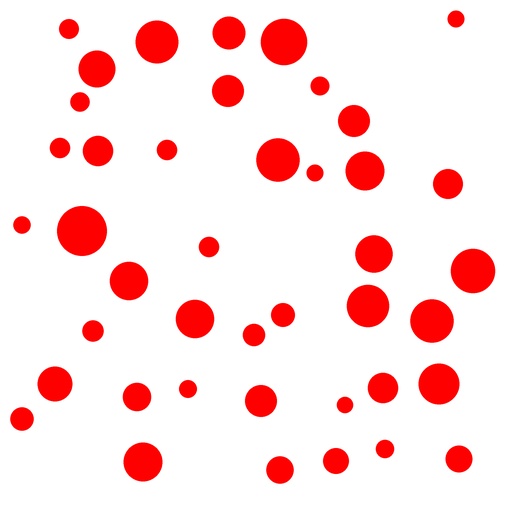}
        \caption{red}
    \end{subfigure}\hfill
    \begin{subfigure}[b]{0.11\textwidth}
        \centering
        \includegraphics[width=\linewidth]{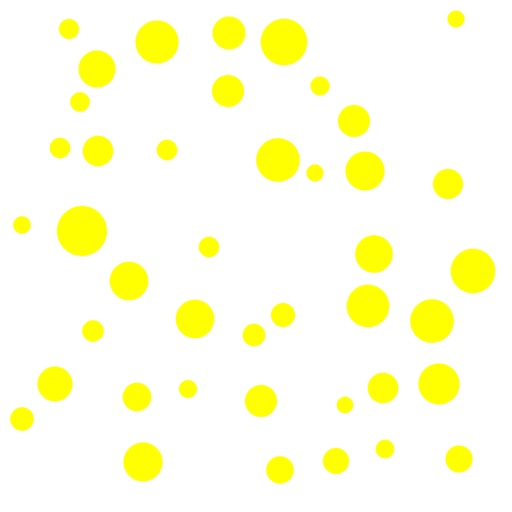}
        \caption{yellow}
    \end{subfigure}
    \vspace{0.3em}
    \begin{subfigure}[b]{0.11\textwidth}
        \centering
        \includegraphics[width=\linewidth]{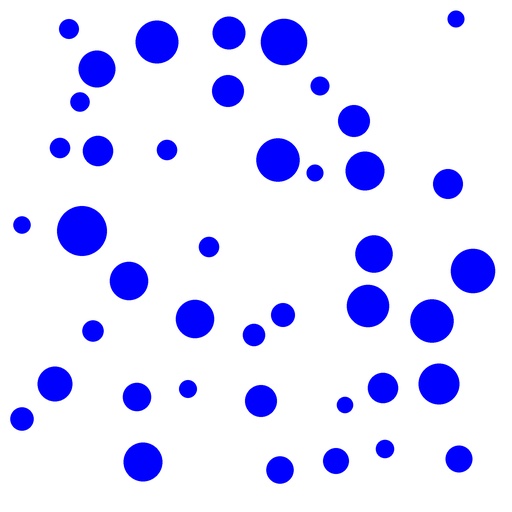}
        \caption{blue}
    \end{subfigure}\hfill
    \begin{subfigure}[b]{0.11\textwidth}
        \centering
        \includegraphics[width=\linewidth]{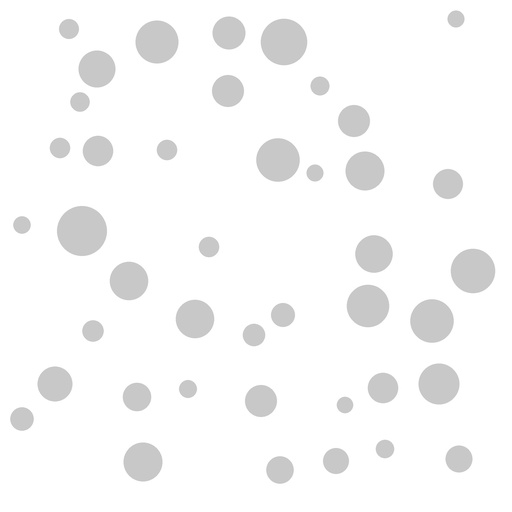}
        \caption{light gray}
    \end{subfigure}\hfill
    \begin{subfigure}[b]{0.11\textwidth}
        \centering
        \includegraphics[width=\linewidth]{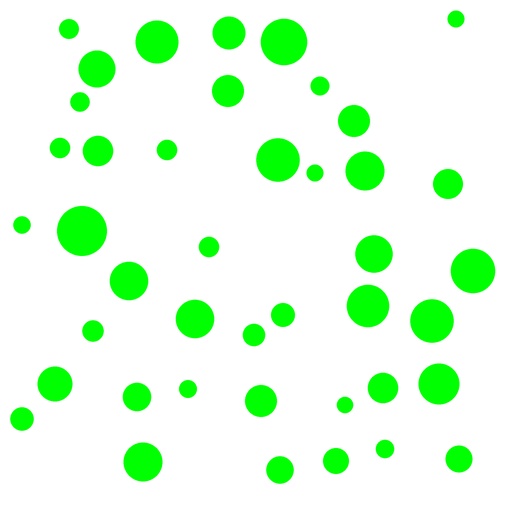}
        \caption{green}
    \end{subfigure}\hfill
    \begin{subfigure}[b]{0.11\textwidth}
        \centering
        \includegraphics[width=\linewidth]{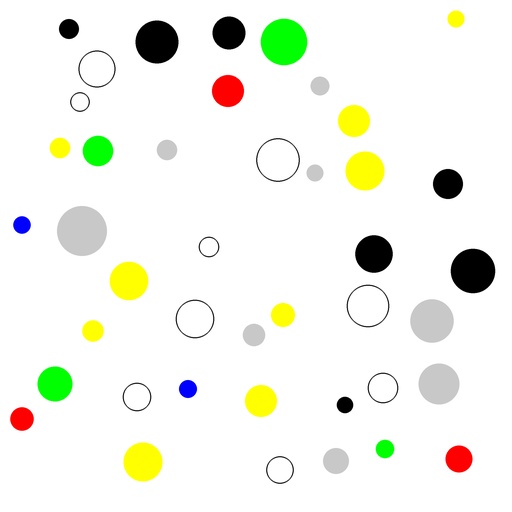}
        \caption{multicolor}
    \end{subfigure}

    \caption{Example images for the \textbf{Object} category, \textbf{Color} pattern, showing different object colors.}
    \label{fig:obj_color_imgs}
\end{figure}

\begin{figure}[!h]
    \centering
    \begin{subfigure}[t]{0.15\textwidth}
        \centering
        \includegraphics[width=\linewidth]{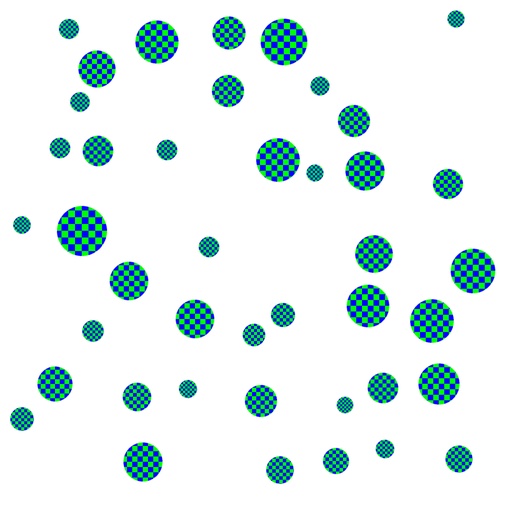}
        \caption{\protect\mbox{Checkerboard}}
    \end{subfigure}\hfill
    \begin{subfigure}[t]{0.15\textwidth}
        \centering
        \includegraphics[width=\linewidth]{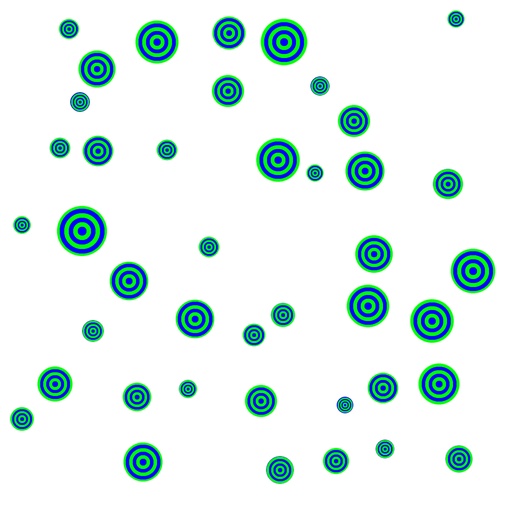}
        \caption{\protect\mbox{Concentric Circles}}
    \end{subfigure}\hfill
    \begin{subfigure}[t]{0.15\textwidth}
        \centering
        \includegraphics[width=\linewidth]{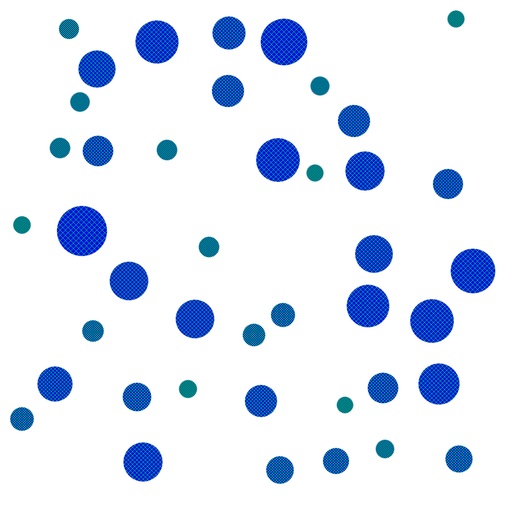}
        \caption{Crosshatch}
    \end{subfigure}

    \vspace{0.3em}
    \begin{subfigure}[t]{0.15\textwidth}
        \centering
        \includegraphics[width=\linewidth]{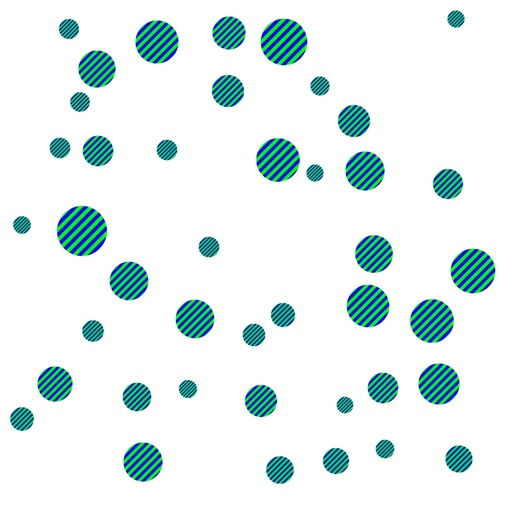}
        \caption{\protect\mbox{Diagonal Stripes}}
    \end{subfigure}\hfill
    \begin{subfigure}[t]{0.15\textwidth}
        \centering
        \includegraphics[width=\linewidth]{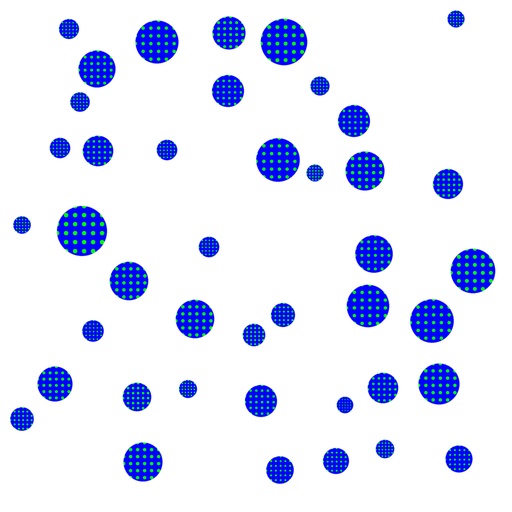}
        \caption{Dots}
    \end{subfigure}\hfill
    \begin{subfigure}[t]{0.15\textwidth}
        \centering
        \includegraphics[width=\linewidth]{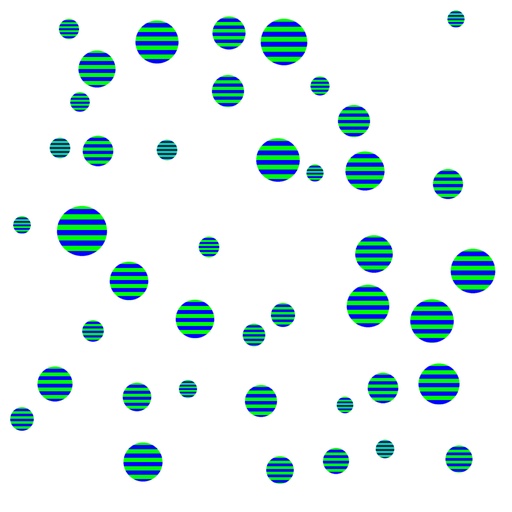}
        \caption{\protect\mbox{Horizontal Stripes}}
    \end{subfigure}

    \vspace{0.3em}
    \begin{subfigure}[t]{0.15\textwidth}
        \centering
        \includegraphics[width=\linewidth]{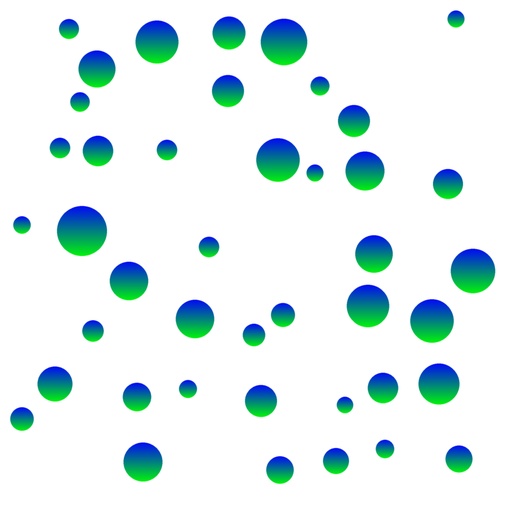}
        \caption{\protect\mbox{Linear Gradient}}
    \end{subfigure}\hfill
    \begin{subfigure}[t]{0.15\textwidth}
        \centering
        \includegraphics[width=\linewidth]{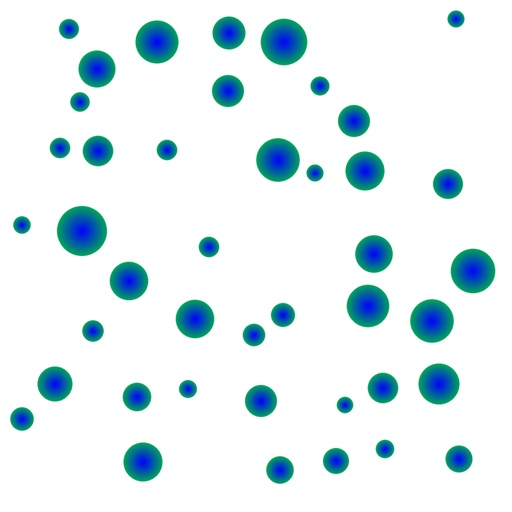}
        \caption{\protect\mbox{Radial Gradient}}
    \end{subfigure}\hfill
    \begin{subfigure}[t]{0.15\textwidth}
        \centering
        \includegraphics[width=\linewidth]{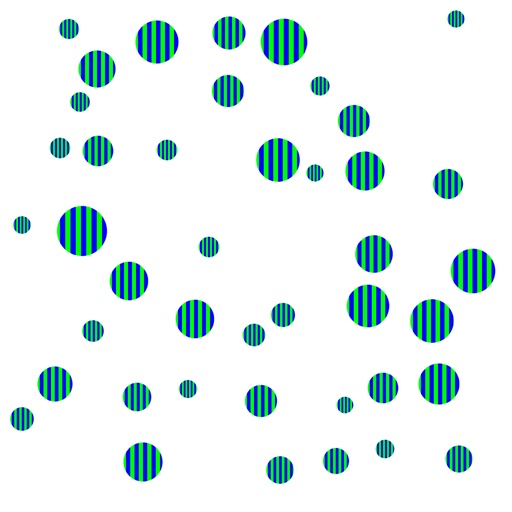}
        \caption{\protect\mbox{Vertical Stripes}}
    \end{subfigure}

    \vspace{0.3em}
    \begin{subfigure}[t]{0.15\textwidth}
        \centering
        \includegraphics[height=2cm]{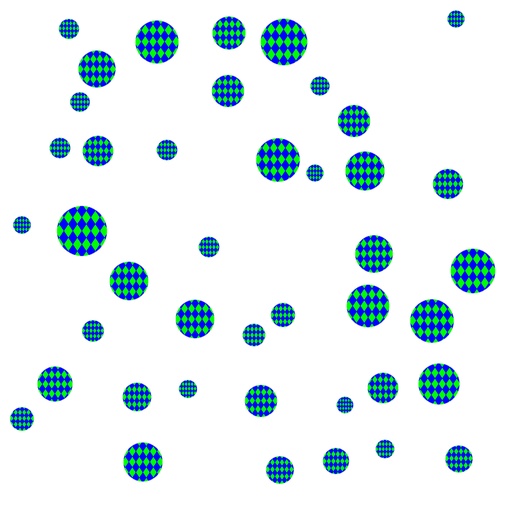}
        \caption{Zigzag}
    \end{subfigure}

    \caption{Example images for the \textbf{Object} category, \textbf{Texture} pattern, showing various texture types.}
    \label{fig:obj_texture_imgs}
\end{figure}

\begin{figure}[H]
    \centering
    \begin{subfigure}[b]{0.14\textwidth}
        \centering
        \includegraphics[width=\linewidth]{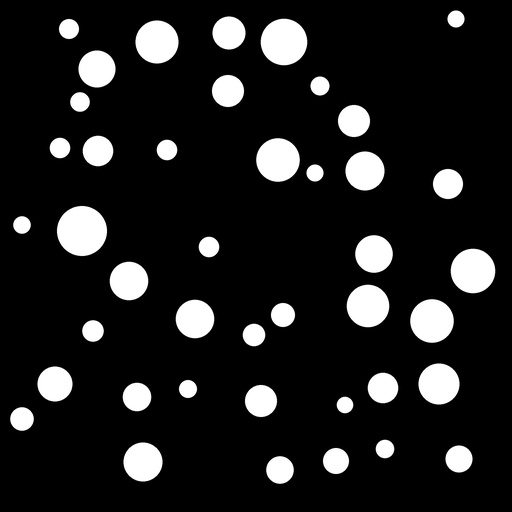}
        \caption{black}
    \end{subfigure}\hfill
    \begin{subfigure}[b]{0.14\textwidth}
        \centering
        \includegraphics[width=\linewidth]{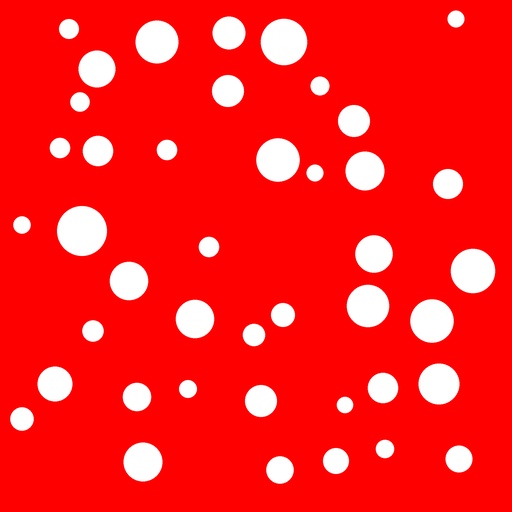}
        \caption{red}
    \end{subfigure}\hfill
    \begin{subfigure}[b]{0.14\textwidth}
        \centering
        \includegraphics[width=\linewidth]{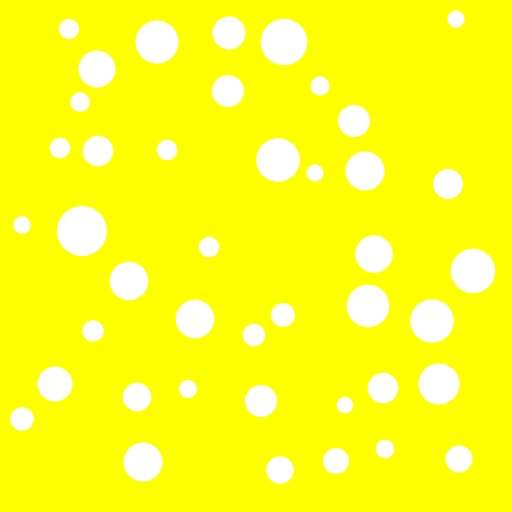}
        \caption{yellow}
    \end{subfigure}
    \vspace{0.3em}
    \begin{subfigure}[b]{0.14\textwidth}
        \centering
        \includegraphics[width=\linewidth]{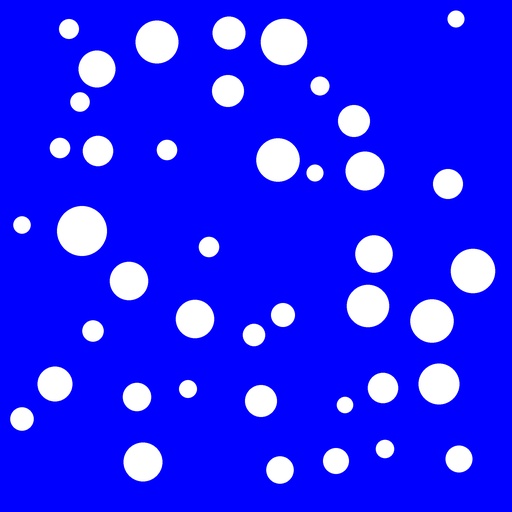}
        \caption{blue}
    \end{subfigure}\hfill
    \begin{subfigure}[b]{0.14\textwidth}
        \centering
        \includegraphics[width=\linewidth]{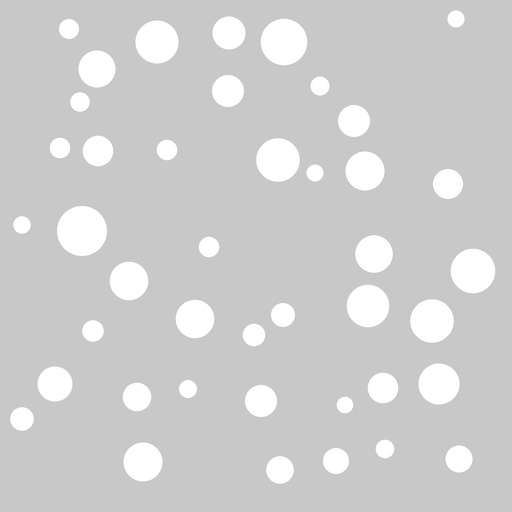}
        \caption{light gray}
    \end{subfigure}\hfill
    \begin{subfigure}[b]{0.14\textwidth}
        \centering
        \includegraphics[width=\linewidth]{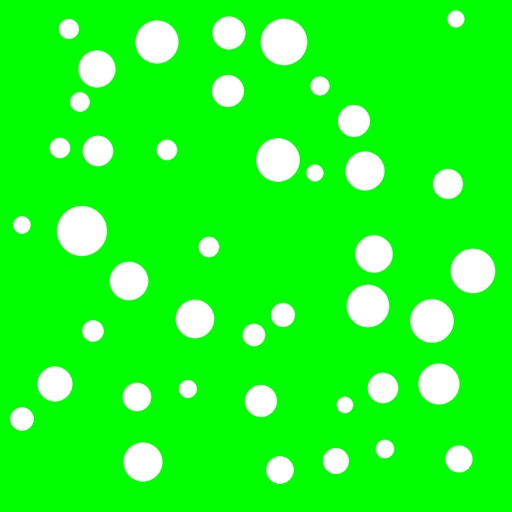}
        \caption{green}
    \end{subfigure}

    \caption{Example images for the \textbf{Background} category, \textbf{Color} pattern, showing different background colors.}
    \label{fig:bg_color_imgs}
\end{figure}

\begin{figure}[H]
    \centering

    \begin{subfigure}[b]{0.14\textwidth}
        \centering
        \includegraphics[width=\linewidth]{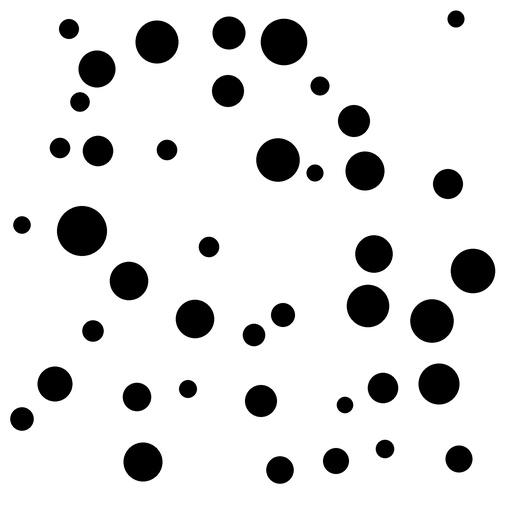}
        \caption{circle}
    \end{subfigure}\hfill
    \begin{subfigure}[b]{0.14\textwidth}
        \centering
        \includegraphics[width=\linewidth]{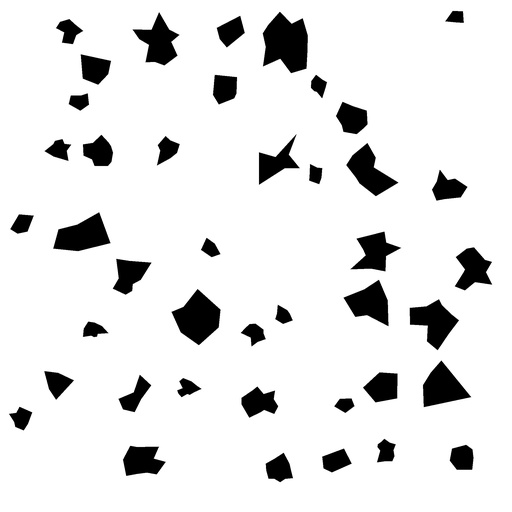}
        \caption{polygon}
    \end{subfigure}\hfill
    \begin{subfigure}[b]{0.14\textwidth}
        \centering
        \includegraphics[width=\linewidth]{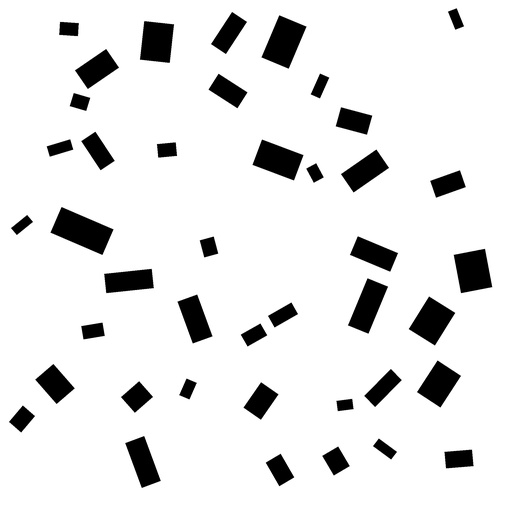}
        \caption{rectangle}
    \end{subfigure}\hfill
    \begin{subfigure}[b]{0.14\textwidth}
        \centering
        \includegraphics[width=\linewidth]{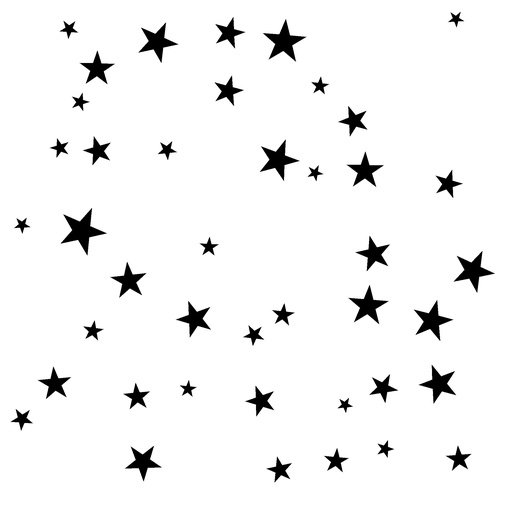}
        \caption{star}
    \end{subfigure}\hfill
    \begin{subfigure}[b]{0.14\textwidth}
        \centering
        \includegraphics[width=\linewidth]{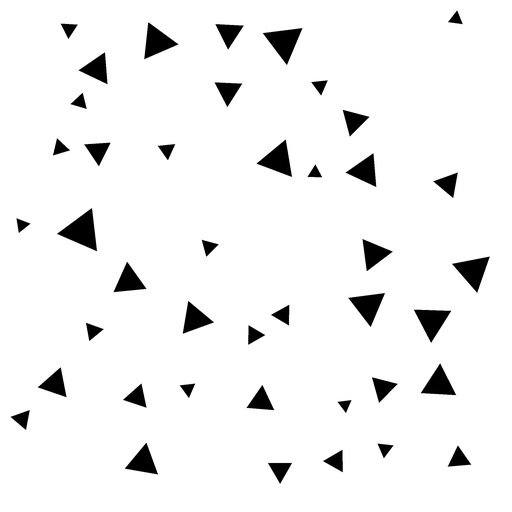}
        \caption{triangle}
    \end{subfigure}

    \caption{Example images for the \textbf{Object} category, \textbf{Shape} pattern, showing different object shapes.}
    \label{fig:obj_shape_imgs}
\end{figure}

\section{Prompts}


\begin{table}[h]
\centering
\small 
\caption{Prompts used when image has different Object Color or Shape.}
\label{tab:texture_prompt_ladder_compact_sup_1}
\begin{tabularx}{\columnwidth}{l >{\raggedright\arraybackslash}p{0.4\columnwidth} >{\raggedright\arraybackslash}p{0.4\columnwidth}}
\toprule
\textbf{ID} & \textbf{Example Prompt Text} & \textbf{Logical Role / Cognitive Cue} \\
\midrule
P1 & \texttt{Count the number of distinct objects in this image...} & \textbf{Baseline:} Generic unconstrained prompt. \\  
\addlinespace
P2 & \texttt{Count the number of \{color\} color objects in this image...} & \textbf{Single (Simple) Attribute:} Simple target Cue (Color) - Replace \{color\} with object colors (blue, green, yellow, gray, multicolor, e.g.). \\
\addlinespace
P3 & \texttt{Count the number of \{color\} color \{shape\} in this image...} & \textbf{Compositional (Simple) Attribute:} Bind target cues (color and shape). Replace \{shape\} with ``circles"(as default in color experiment), ``squares", ``triangles",``stars", etc. for shape experiment.  Replace \{color\} with ``black"(as default in shape experiment), ``yellow", ``red",``blue", etc. for color experiment.\\\\ 
\bottomrule
\end{tabularx}
\end{table}

\begin{table}[h]
\centering
\small 
\caption{Prompts used when image has different Background Texture.}
\label{tab:bgtexture_prompt_ladder_compact}
\begin{tabularx}{\columnwidth}{l >{\raggedright\arraybackslash}p{0.4\columnwidth} >{\raggedright\arraybackslash}p{0.4\columnwidth}}
\toprule
\textbf{ID} & \textbf{Example Prompt Text} & \textbf{Logical Role / Cognitive Cue} \\
\midrule
P1 & \texttt{Count the number of distinct objects in this image...} & \textbf{Baseline:} Generic unconstrained prompt.   \\
\addlinespace
P2 & \texttt{Count the number of \{color\} color objects in this image...} & \textbf{Single (Simple) Attribute:} Simple Object Cue (Color) - Replace \{color\} with object colors (``white" for default). \\
\addlinespace
P3 & \texttt{Count the number of \{color\} color \{shape\} in this image...}  & \textbf{Compositional (Target):}  Binding (Simple+ Simple). Tests binding between two independent object attributes - Replace \{shape\} with object shape (``circles" for default).\\\\
\addlinespace
P4 & \texttt{Count the number of \{color\} color objects in this image with \{pattern\} background...} & \textbf{Compositional (Target+):}  Binding (Complex + Simple). Tests binding a simple cue with a complex one. Tests whether the model can integrate object-level and background-level features.\\
\addlinespace
P5 &  \texttt{Count the number of \{color\} color {shape} in this image with \{color\}\{pattern\} background...} & \textbf{Compositional (High Load):} Multi-attribute binding under high cognitive load.  object color + background color (``blue-green" for default) + background pattern\\
\bottomrule
\end{tabularx}
\end{table}

\begin{table}[h]
\centering
\small 
\caption{Prompts used when image has different Background Color.}
\label{tab:bgcolor_prompt_ladder_compact}
\begin{tabularx}{\columnwidth}{l >{\raggedright\arraybackslash}p{0.4\columnwidth} >{\raggedright\arraybackslash}p{0.4\columnwidth}}
\toprule
\textbf{ID} & \textbf{Example Prompt Text} & \textbf{Logical Role / Cognitive Cue} \\
\midrule
P1 & \texttt{Count the number of distinct objects in this image...} & \textbf{Baseline:} Generic unconstrained prompt.   \\
\addlinespace
P2 & \texttt{Count the number of \{color\} objects in this image with \{color\} background...} & \textbf{Compositional (Simple) Attribute:} Bind target cues and contextual cue: Tests selective filtering based on a single attribute (background color), with object color( "white" for default). \\
\addlinespace
P3 & \texttt{Count the number of \{color\} \{shape\} in this image with \{color\}\ background...} & \textbf{Compositional (Complex) Attribute:} Bind target and contextual cue: a single background attribute (background color), with two object attributes( "white circle" for default). \\\\
\bottomrule
\end{tabularx}
\end{table}

\section{Effects of Visual Complexity}
Tables \ref{tab:error_texture_prompt1}, Table \ref{tab:error_texture_prompt3}, Table \ref{tab:error_texture_prompt4}, and Table \ref{tab:error_texture_prompt5} present the Mean Relative Count Error (MRCE) for prompts 1, 3, 4, and 5, respectively.

\begin{table}[h]
\caption{Mean Relative Count Error (lower is better) for Prompt 1 across all patterns. ``Bg'' denotes \textit{Background}, ``Obj'' denotes \textit{Object} and ''diag. str.'' denotes diagonal stripes, ''ver. str.'' denotes vertical stripes, ''hor. str.'' denotes horizontal stripes, ''con. cir.'' denotes concentric circles, ''lin. grad.'' linear gradient, ''rad. grad.'' denotes radial gradient, ''con. rgs.'' denotes concentric rings, ''cr. hatch'' denotes cross hatch categories.}
\label{tab:error_texture_prompt1}
\renewcommand{\arraystretch}{0.85} 
\setlength{\tabcolsep}{3pt}        
\centering
\begin{tabular}{@{}llp{1.5cm}cccc@{}}
\toprule
Cat. & Feat. & Pattern & Qwen7b & Qwen32b & Intern & Kimi \\
\midrule
Bg & Color & blue & \cellcolor{blue!60}{\textcolor{black}{0.154}} & \cellcolor{blue!56}{\textcolor{black}{0.142}} & \cellcolor{blue!58}{\textcolor{black}{0.103}} & \cellcolor{blue!58}{\textcolor{black}{0.079}} \\
Bg & Color & black & \cellcolor{blue!58}{\textcolor{black}{0.170}} & \cellcolor{blue!55}{\textcolor{black}{0.143}} & \cellcolor{blue!57}{\textcolor{black}{0.115}} & \cellcolor{blue!58}{\textcolor{black}{0.080}} \\
Bg & Color & green & \cellcolor{blue!53}{\textcolor{black}{0.235}} & \cellcolor{blue!57}{\textcolor{black}{0.112}} & \cellcolor{blue!57}{\textcolor{black}{0.112}} & \cellcolor{blue!53}{\textcolor{black}{0.145}} \\
Bg & Color & gray & \cellcolor{blue!51}{\textcolor{black}{0.254}} & \cellcolor{blue!56}{\textcolor{black}{0.135}} & \cellcolor{blue!56}{\textcolor{black}{0.138}} & \cellcolor{blue!58}{\textcolor{black}{0.078}} \\
Bg & Color & red & \cellcolor{blue!47}{\textcolor{black}{0.311}} & \cellcolor{blue!56}{\textcolor{black}{0.129}} & \cellcolor{blue!56}{\textcolor{black}{0.131}} & \cellcolor{blue!45}{\textcolor{black}{0.247}} \\
Bg & Color & yellow & \cellcolor{blue!46}{\textcolor{black}{0.318}} & \cellcolor{blue!56}{\textcolor{black}{0.138}} & \cellcolor{blue!53}{\textcolor{black}{0.198}} & \cellcolor{blue!38}{\textcolor{black}{0.341}} \\
Bg & Texture & noise & \cellcolor{blue!53}{\textcolor{black}{0.234}} & \cellcolor{blue!57}{\textcolor{black}{0.121}} & \cellcolor{blue!56}{\textcolor{black}{0.128}} & \cellcolor{blue!58}{\textcolor{black}{0.086}} \\
Bg & Texture & cr. hatch & \cellcolor{blue!43}{\textcolor{black}{0.355}} & \cellcolor{blue!56}{\textcolor{black}{0.141}} & \cellcolor{blue!57}{\textcolor{black}{0.122}} & \cellcolor{blue!58}{\textcolor{black}{0.078}} \\
Bg & Texture & lin. grad. & \cellcolor{blue!31}{\textcolor{black}{0.504}} & \cellcolor{blue!57}{\textcolor{black}{0.115}} & \cellcolor{blue!59}{\textcolor{black}{0.078}} & \cellcolor{blue!58}{\textcolor{black}{0.084}} \\
Bg & Texture & rad. grad. & \cellcolor{blue!25}{\textcolor{black}{0.572}} & \cellcolor{blue!57}{\textcolor{black}{0.120}} & \cellcolor{blue!58}{\textcolor{black}{0.100}} & \cellcolor{blue!54}{\textcolor{black}{0.139}} \\
Bg & Texture & checkerboard & \cellcolor{blue!10}{\textcolor{black}{0.832}} & \cellcolor{blue!56}{\textcolor{black}{0.141}} & \cellcolor{blue!55}{\textcolor{black}{0.160}} & \cellcolor{blue!31}{\textcolor{black}{0.432}} \\
Bg & Texture & dots & \cellcolor{blue!10}{\textcolor{black}{0.803}} & \cellcolor{blue!53}{\textcolor{black}{0.184}} & \cellcolor{blue!56}{\textcolor{black}{0.144}} & \cellcolor{blue!30}{\textcolor{black}{0.453}} \\
Bg & Texture & diag. str. & \cellcolor{blue!11}{\textcolor{black}{0.751}} & \cellcolor{blue!54}{\textcolor{black}{0.163}} & \cellcolor{blue!58}{\textcolor{black}{0.098}} & \cellcolor{blue!12}{\textcolor{black}{0.687}} \\
Bg & Texture & con. rgs & \cellcolor{blue!13}{\textcolor{black}{0.723}} & \cellcolor{blue!49}{\textcolor{black}{0.270}} & \cellcolor{blue!53}{\textcolor{black}{0.198}} & \cellcolor{blue!16}{\textcolor{black}{0.627}} \\
Bg & Texture & hor. str. & \cellcolor{blue!12}{\textcolor{black}{0.734}} & \cellcolor{blue!47}{\textcolor{black}{0.306}} & \cellcolor{blue!57}{\textcolor{black}{0.109}} & \cellcolor{blue!10}{\textcolor{black}{0.757}} \\
Bg & Texture & ver. str. & \cellcolor{blue!15}{\textcolor{black}{0.695}} & \cellcolor{blue!41}{\textcolor{black}{0.413}} & \cellcolor{blue!55}{\textcolor{black}{0.153}} & \cellcolor{blue!10}{\textcolor{black}{0.773}} \\
Bg & Texture & bubbles & \cellcolor{blue!10}{\textcolor{black}{0.888}} & \cellcolor{blue!10}{\textcolor{black}{1.209}} & \cellcolor{blue!10}{\textcolor{black}{1.203}} & \cellcolor{blue!10}{\textcolor{black}{0.845}} \\
Obj & Color & white & \cellcolor{blue!55}{\textcolor{black}{0.210}} & \cellcolor{blue!60}{\textcolor{black}{0.066}} & \cellcolor{blue!58}{\textcolor{black}{0.102}} & \cellcolor{blue!57}{\textcolor{black}{0.098}} \\
Obj & Color & red & \cellcolor{blue!57}{\textcolor{black}{0.187}} & \cellcolor{blue!56}{\textcolor{black}{0.130}} & \cellcolor{blue!59}{\textcolor{black}{0.084}} & \cellcolor{blue!58}{\textcolor{black}{0.084}} \\
Obj & Color & yellow & \cellcolor{blue!56}{\textcolor{black}{0.201}} & \cellcolor{blue!57}{\textcolor{black}{0.113}} & \cellcolor{blue!59}{\textcolor{black}{0.082}} & \cellcolor{blue!55}{\textcolor{black}{0.116}} \\
Obj & Color & blue & \cellcolor{blue!54}{\textcolor{black}{0.217}} & \cellcolor{blue!56}{\textcolor{black}{0.130}} & \cellcolor{blue!57}{\textcolor{black}{0.108}} & \cellcolor{blue!59}{\textcolor{black}{0.068}} \\
Obj & Color & light gray & \cellcolor{blue!55}{\textcolor{black}{0.215}} & \cellcolor{blue!58}{\textcolor{black}{0.088}} & \cellcolor{blue!55}{\textcolor{black}{0.158}} & \cellcolor{blue!57}{\textcolor{black}{0.097}} \\
Obj & Color & green & \cellcolor{blue!45}{\textcolor{black}{0.335}} & \cellcolor{blue!59}{\textcolor{black}{0.075}} & \cellcolor{blue!59}{\textcolor{black}{0.081}} & \cellcolor{blue!58}{\textcolor{black}{0.078}} \\
Obj & Color & multicolor & \cellcolor{blue!27}{\textcolor{black}{0.553}} & \cellcolor{blue!56}{\textcolor{black}{0.129}} & \cellcolor{blue!59}{\textcolor{black}{0.081}} & \cellcolor{blue!37}{\textcolor{black}{0.350}} \\
Obj & Shape & star & \cellcolor{blue!54}{\textcolor{black}{0.216}} & \cellcolor{blue!55}{\textcolor{black}{0.143}} & \cellcolor{blue!59}{\textcolor{black}{0.083}} & \cellcolor{blue!58}{\textcolor{black}{0.077}} \\
Obj & Shape & circle & \cellcolor{blue!58}{\textcolor{black}{0.178}} & \cellcolor{blue!56}{\textcolor{black}{0.137}} & \cellcolor{blue!55}{\textcolor{black}{0.154}} & \cellcolor{blue!59}{\textcolor{black}{0.072}} \\
Obj & Shape & rectangle & \cellcolor{blue!40}{\textcolor{black}{0.390}} & \cellcolor{blue!57}{\textcolor{black}{0.123}} & \cellcolor{blue!53}{\textcolor{black}{0.195}} & \cellcolor{blue!58}{\textcolor{black}{0.085}} \\
Obj & Shape & polygon & \cellcolor{blue!40}{\textcolor{black}{0.397}} & \cellcolor{blue!57}{\textcolor{black}{0.118}} & \cellcolor{blue!56}{\textcolor{black}{0.140}} & \cellcolor{blue!54}{\textcolor{black}{0.138}} \\
Obj & Shape & triangle & \cellcolor{blue!32}{\textcolor{black}{0.493}} & \cellcolor{blue!55}{\textcolor{black}{0.154}} & \cellcolor{blue!60}{\textcolor{black}{0.069}} & \cellcolor{blue!40}{\textcolor{black}{0.320}} \\
Obj & Texture & rad. grad. & \cellcolor{blue!56}{\textcolor{black}{0.200}} & \cellcolor{blue!56}{\textcolor{black}{0.124}} & \cellcolor{blue!59}{\textcolor{black}{0.085}} & \cellcolor{blue!59}{\textcolor{black}{0.074}} \\
Obj & Texture & dots & \cellcolor{blue!34}{\textcolor{black}{0.462}} & \cellcolor{blue!55}{\textcolor{black}{0.143}} & \cellcolor{blue!56}{\textcolor{black}{0.132}} & \cellcolor{blue!59}{\textcolor{black}{0.071}} \\
Obj & Texture & con. cir. & \cellcolor{blue!25}{\textcolor{black}{0.576}} & \cellcolor{blue!59}{\textcolor{black}{0.075}} & \cellcolor{blue!58}{\textcolor{black}{0.106}} & \cellcolor{blue!59}{\textcolor{black}{0.073}} \\
Obj & Texture & lin. grad. & \cellcolor{blue!25}{\textcolor{black}{0.576}} & \cellcolor{blue!57}{\textcolor{black}{0.113}} & \cellcolor{blue!59}{\textcolor{black}{0.077}} & \cellcolor{blue!59}{\textcolor{black}{0.067}} \\
Obj & Texture & cr. hatch & \cellcolor{blue!22}{\textcolor{black}{0.617}} & \cellcolor{blue!59}{\textcolor{black}{0.076}} & \cellcolor{blue!58}{\textcolor{black}{0.101}} & \cellcolor{blue!58}{\textcolor{black}{0.079}} \\
Obj & Texture & checkerboard & \cellcolor{blue!19}{\textcolor{black}{0.644}} & \cellcolor{blue!57}{\textcolor{black}{0.110}} & \cellcolor{blue!58}{\textcolor{black}{0.101}} & \cellcolor{blue!60}{\textcolor{black}{0.062}} \\
Obj & Texture & ver. str. & \cellcolor{blue!22}{\textcolor{black}{0.617}} & \cellcolor{blue!56}{\textcolor{black}{0.128}} & \cellcolor{blue!58}{\textcolor{black}{0.089}} & \cellcolor{blue!57}{\textcolor{black}{0.092}} \\
Obj & Texture & zigzag & \cellcolor{blue!27}{\textcolor{black}{0.551}} & \cellcolor{blue!55}{\textcolor{black}{0.148}} & \cellcolor{blue!55}{\textcolor{black}{0.153}} & \cellcolor{blue!58}{\textcolor{black}{0.079}} \\
Obj & Texture & diag. str. & \cellcolor{blue!27}{\textcolor{black}{0.548}} & \cellcolor{blue!53}{\textcolor{black}{0.198}} & \cellcolor{blue!57}{\textcolor{black}{0.125}} & \cellcolor{blue!59}{\textcolor{black}{0.068}} \\
Obj & Texture & hor. str. & \cellcolor{blue!23}{\textcolor{black}{0.605}} & \cellcolor{blue!45}{\textcolor{black}{0.342}} & \cellcolor{blue!55}{\textcolor{black}{0.156}} & \cellcolor{blue!57}{\textcolor{black}{0.089}} \\
\bottomrule
\end{tabular}
\end{table}

\begin{table}[h]
\caption{Mean Relative Count Error (lower is better) for Prompt 3 across all patterns. ``Bg'' denotes \textit{Background}, ``Obj'' denotes \textit{Object} and ''diag. str.'' denotes diagonal stripes, ''ver. str.'' denotes vertical stripes, ''hor. str.'' denotes horizontal stripes, ''con. cir.'' denotes concentric circles, ''lin. grad.'' linear gradient, ''rad. grad.'' denotes radial gradient, ''con. rgs.'' denotes concentric rings, ''cr. hatch'' denotes cross hatch categories.}
\label{tab:error_texture_prompt3}
\renewcommand{\arraystretch}{0.85} 
\setlength{\tabcolsep}{3pt}        
\centering

\begin{tabular}{@{}llp{1.5cm}cccc@{}}
\toprule
Cat. & Feat. & Pattern & Qwen7b & Qwen32b & Intern & Kimi \\
\midrule
Bg & Color & blue & \cellcolor{blue!53}{\textcolor{black}{0.142}} & \cellcolor{blue!54}{\textcolor{black}{0.096}} & \cellcolor{blue!55}{\textcolor{black}{0.130}} & \cellcolor{blue!59}{\textcolor{black}{0.064}} \\
Bg & Color & green & \cellcolor{blue!50}{\textcolor{black}{0.164}} & \cellcolor{blue!55}{\textcolor{black}{0.092}} & \cellcolor{blue!54}{\textcolor{black}{0.143}} & \cellcolor{blue!58}{\textcolor{black}{0.070}} \\
Bg & Color & black & \cellcolor{blue!50}{\textcolor{black}{0.168}} & \cellcolor{blue!56}{\textcolor{black}{0.085}} & \cellcolor{blue!54}{\textcolor{black}{0.141}} & \cellcolor{blue!55}{\textcolor{black}{0.079}} \\
Bg & Color & red & \cellcolor{blue!50}{\textcolor{black}{0.166}} & \cellcolor{blue!54}{\textcolor{black}{0.099}} & \cellcolor{blue!51}{\textcolor{black}{0.176}} & \cellcolor{blue!58}{\textcolor{black}{0.067}} \\
Bg & Color & gray & \cellcolor{blue!47}{\textcolor{black}{0.186}} & \cellcolor{blue!53}{\textcolor{black}{0.106}} & \cellcolor{blue!53}{\textcolor{black}{0.159}} & \cellcolor{blue!56}{\textcolor{black}{0.076}} \\
Bg & Color & yellow & \cellcolor{blue!46}{\textcolor{black}{0.190}} & \cellcolor{blue!54}{\textcolor{black}{0.096}} & \cellcolor{blue!48}{\textcolor{black}{0.216}} & \cellcolor{blue!52}{\textcolor{black}{0.094}} \\
Bg & Texture & lin. grad. & \cellcolor{blue!50}{\textcolor{black}{0.164}} & \cellcolor{blue!55}{\textcolor{black}{0.093}} & \cellcolor{blue!60}{\textcolor{black}{0.072}} & \cellcolor{blue!51}{\textcolor{black}{0.098}} \\
Bg & Texture & noise & \cellcolor{blue!53}{\textcolor{black}{0.144}} & \cellcolor{blue!55}{\textcolor{black}{0.092}} & \cellcolor{blue!52}{\textcolor{black}{0.161}} & \cellcolor{blue!57}{\textcolor{black}{0.073}} \\
Bg & Texture & rad. grad. & \cellcolor{blue!48}{\textcolor{black}{0.179}} & \cellcolor{blue!49}{\textcolor{black}{0.138}} & \cellcolor{blue!54}{\textcolor{black}{0.137}} & \cellcolor{blue!56}{\textcolor{black}{0.078}} \\
Bg & Texture & ver. str. & \cellcolor{blue!41}{\textcolor{black}{0.224}} & \cellcolor{blue!42}{\textcolor{black}{0.183}} & \cellcolor{blue!49}{\textcolor{black}{0.203}} & \cellcolor{blue!56}{\textcolor{black}{0.077}} \\
Bg & Texture & checkerboard & \cellcolor{blue!38}{\textcolor{black}{0.243}} & \cellcolor{blue!45}{\textcolor{black}{0.166}} & \cellcolor{blue!50}{\textcolor{black}{0.187}} & \cellcolor{blue!45}{\textcolor{black}{0.121}} \\
Bg & Texture & dots & \cellcolor{blue!33}{\textcolor{black}{0.280}} & \cellcolor{blue!42}{\textcolor{black}{0.183}} & \cellcolor{blue!51}{\textcolor{black}{0.173}} & \cellcolor{blue!49}{\textcolor{black}{0.107}} \\
Bg & Texture & hor. str. & \cellcolor{blue!38}{\textcolor{black}{0.246}} & \cellcolor{blue!37}{\textcolor{black}{0.219}} & \cellcolor{blue!55}{\textcolor{black}{0.126}} & \cellcolor{blue!34}{\textcolor{black}{0.167}} \\
Bg & Texture & con. rgs & \cellcolor{blue!37}{\textcolor{black}{0.254}} & \cellcolor{blue!42}{\textcolor{black}{0.186}} & \cellcolor{blue!44}{\textcolor{black}{0.261}} & \cellcolor{blue!58}{\textcolor{black}{0.070}} \\
Bg & Texture & cr. hatch & \cellcolor{blue!45}{\textcolor{black}{0.201}} & \cellcolor{blue!10}{\textcolor{black}{0.495}} & \cellcolor{blue!54}{\textcolor{black}{0.140}} & \cellcolor{blue!58}{\textcolor{black}{0.067}} \\
Bg & Texture & diag. str. & \cellcolor{blue!36}{\textcolor{black}{0.260}} & \cellcolor{blue!11}{\textcolor{black}{0.411}} & \cellcolor{blue!52}{\textcolor{black}{0.162}} & \cellcolor{blue!55}{\textcolor{black}{0.082}} \\
Bg & Texture & bubbles & \cellcolor{blue!34}{\textcolor{black}{0.271}} & \cellcolor{blue!41}{\textcolor{black}{0.196}} & \cellcolor{blue!38}{\textcolor{black}{0.344}} & \cellcolor{blue!25}{\textcolor{black}{0.205}} \\
Obj & Color & green & \cellcolor{blue!59}{\textcolor{black}{0.102}} & \cellcolor{blue!59}{\textcolor{black}{0.064}} & \cellcolor{blue!59}{\textcolor{black}{0.074}} & \cellcolor{blue!56}{\textcolor{black}{0.077}} \\
Obj & Color & blue & \cellcolor{blue!59}{\textcolor{black}{0.105}} & \cellcolor{blue!55}{\textcolor{black}{0.092}} & \cellcolor{blue!57}{\textcolor{black}{0.106}} & \cellcolor{blue!60}{\textcolor{black}{0.062}} \\
Obj & Color & red & \cellcolor{blue!56}{\textcolor{black}{0.126}} & \cellcolor{blue!53}{\textcolor{black}{0.108}} & \cellcolor{blue!58}{\textcolor{black}{0.088}} & \cellcolor{blue!58}{\textcolor{black}{0.067}} \\
Obj & Color & yellow & \cellcolor{blue!53}{\textcolor{black}{0.142}} & \cellcolor{blue!60}{\textcolor{black}{0.058}} & \cellcolor{blue!58}{\textcolor{black}{0.093}} & \cellcolor{blue!49}{\textcolor{black}{0.108}} \\
Obj & Color & white & \cellcolor{blue!58}{\textcolor{black}{0.111}} & \cellcolor{blue!50}{\textcolor{black}{0.125}} & \cellcolor{blue!55}{\textcolor{black}{0.128}} & \cellcolor{blue!43}{\textcolor{black}{0.130}} \\
Obj & Color & light gray & \cellcolor{blue!48}{\textcolor{black}{0.176}} & \cellcolor{blue!52}{\textcolor{black}{0.113}} & \cellcolor{blue!50}{\textcolor{black}{0.188}} & \cellcolor{blue!54}{\textcolor{black}{0.085}} \\
Obj & Color & multicolor & \cellcolor{blue!27}{\textcolor{black}{0.319}} & \cellcolor{blue!29}{\textcolor{black}{0.279}} & \cellcolor{blue!44}{\textcolor{black}{0.259}} & \cellcolor{blue!10}{\textcolor{black}{0.313}} \\
Obj & Shape & polygon & \cellcolor{blue!58}{\textcolor{black}{0.112}} & \cellcolor{blue!56}{\textcolor{black}{0.084}} & \cellcolor{blue!53}{\textcolor{black}{0.148}} & \cellcolor{blue!55}{\textcolor{black}{0.082}} \\
Obj & Shape & star & \cellcolor{blue!53}{\textcolor{black}{0.148}} & \cellcolor{blue!48}{\textcolor{black}{0.144}} & \cellcolor{blue!58}{\textcolor{black}{0.090}} & \cellcolor{blue!54}{\textcolor{black}{0.085}} \\
Obj & Shape & circle & \cellcolor{blue!47}{\textcolor{black}{0.182}} & \cellcolor{blue!49}{\textcolor{black}{0.134}} & \cellcolor{blue!51}{\textcolor{black}{0.182}} & \cellcolor{blue!56}{\textcolor{black}{0.076}} \\
Obj & Shape & triangle & \cellcolor{blue!60}{\textcolor{black}{0.101}} & \cellcolor{blue!47}{\textcolor{black}{0.148}} & \cellcolor{blue!58}{\textcolor{black}{0.088}} & \cellcolor{blue!10}{\textcolor{black}{0.281}} \\
Obj & Shape & rectangle & \cellcolor{blue!49}{\textcolor{black}{0.169}} & \cellcolor{blue!43}{\textcolor{black}{0.176}} & \cellcolor{blue!47}{\textcolor{black}{0.230}} & \cellcolor{blue!55}{\textcolor{black}{0.082}} \\
Obj & Texture & con. cir. & \cellcolor{blue!56}{\textcolor{black}{0.124}} & \cellcolor{blue!53}{\textcolor{black}{0.104}} & \cellcolor{blue!52}{\textcolor{black}{0.170}} & \cellcolor{blue!58}{\textcolor{black}{0.070}} \\
Obj & Texture & ver. str. & \cellcolor{blue!48}{\textcolor{black}{0.179}} & \cellcolor{blue!42}{\textcolor{black}{0.183}} & \cellcolor{blue!52}{\textcolor{black}{0.164}} & \cellcolor{blue!54}{\textcolor{black}{0.084}} \\
Obj & Texture & hor. str. & \cellcolor{blue!46}{\textcolor{black}{0.194}} & \cellcolor{blue!37}{\textcolor{black}{0.219}} & \cellcolor{blue!50}{\textcolor{black}{0.188}} & \cellcolor{blue!54}{\textcolor{black}{0.083}} \\
Obj & Texture & lin. grad. & \cellcolor{blue!48}{\textcolor{black}{0.176}} & \cellcolor{blue!55}{\textcolor{black}{0.093}} & \cellcolor{blue!31}{\textcolor{black}{0.421}} & \cellcolor{blue!56}{\textcolor{black}{0.077}} \\
Obj & Texture & dots & \cellcolor{blue!34}{\textcolor{black}{0.270}} & \cellcolor{blue!42}{\textcolor{black}{0.183}} & \cellcolor{blue!40}{\textcolor{black}{0.310}} & \cellcolor{blue!59}{\textcolor{black}{0.065}} \\
Obj & Texture & rad. grad. & \cellcolor{blue!42}{\textcolor{black}{0.217}} & \cellcolor{blue!49}{\textcolor{black}{0.138}} & \cellcolor{blue!31}{\textcolor{black}{0.432}} & \cellcolor{blue!54}{\textcolor{black}{0.085}} \\
Obj & Texture & checkerboard & \cellcolor{blue!22}{\textcolor{black}{0.355}} & \cellcolor{blue!23}{\textcolor{black}{0.326}} & \cellcolor{blue!26}{\textcolor{black}{0.492}} & \cellcolor{blue!56}{\textcolor{black}{0.078}} \\
Obj & Texture & zigzag & \cellcolor{blue!36}{\textcolor{black}{0.261}} & \cellcolor{blue!17}{\textcolor{black}{0.369}} & \cellcolor{blue!11}{\textcolor{black}{0.677}} & \cellcolor{blue!53}{\textcolor{black}{0.088}} \\
Obj & Texture & diag. str. & \cellcolor{blue!10}{\textcolor{black}{0.504}} & \cellcolor{blue!11}{\textcolor{black}{0.411}} & \cellcolor{blue!22}{\textcolor{black}{0.535}} & \cellcolor{blue!53}{\textcolor{black}{0.091}} \\
Obj & Texture & cr. hatch & \cellcolor{blue!10}{\textcolor{black}{0.490}} & \cellcolor{blue!10}{\textcolor{black}{0.495}} & \cellcolor{blue!10}{\textcolor{black}{0.819}} & \cellcolor{blue!10}{\textcolor{black}{0.301}} \\
\bottomrule
\end{tabular}
\end{table}

\begin{table}[h]
\caption{Mean Relative Count Error (lower is better) for Prompt 4 across all patterns. ``Bg'' denotes \textit{Background}, ``Obj'' denotes \textit{Object} and ''diag. str.'' denotes diagonal stripes, ''ver. str.'' denotes vertical stripes, ''hor. str.'' denotes horizontal stripes, ''con. cir.'' denotes concentric circles, ''lin. grad.'' linear gradient, ''rad. grad.'' denotes radial gradient, ''con. rgs.'' denotes concentric rings, ''cr. hatch'' denotes cross hatch categories.}
\label{tab:error_texture_prompt4}
\renewcommand{\arraystretch}{0.85} 
\setlength{\tabcolsep}{3pt}        
\centering

\begin{tabular}{@{}llp{1.5cm}cccc@{}}
\toprule
Cat. & Feat. & Pattern & Qwen7b & Qwen32b & Intern & Kimi \\
\midrule
Bg & Texture & lin. grad. & \cellcolor{blue!54}{\textcolor{black}{0.183}} & \cellcolor{blue!58}{\textcolor{black}{0.103}} & \cellcolor{blue!60}{\textcolor{black}{0.112}} & \cellcolor{blue!58}{\textcolor{black}{0.079}} \\
Bg & Texture & noise & \cellcolor{blue!60}{\textcolor{black}{0.137}} & \cellcolor{blue!60}{\textcolor{black}{0.089}} & \cellcolor{blue!49}{\textcolor{black}{0.197}} & \cellcolor{blue!58}{\textcolor{black}{0.070}} \\
Bg & Texture & rad. grad. & \cellcolor{blue!54}{\textcolor{black}{0.187}} & \cellcolor{blue!56}{\textcolor{black}{0.138}} & \cellcolor{blue!57}{\textcolor{black}{0.135}} & \cellcolor{blue!59}{\textcolor{black}{0.063}} \\
Bg & Texture & cr. hatch & \cellcolor{blue!53}{\textcolor{black}{0.192}} & \cellcolor{blue!56}{\textcolor{black}{0.138}} & \cellcolor{blue!52}{\textcolor{black}{0.172}} & \cellcolor{blue!56}{\textcolor{black}{0.105}} \\
Bg & Texture & checkerboard & \cellcolor{blue!48}{\textcolor{black}{0.245}} & \cellcolor{blue!57}{\textcolor{black}{0.122}} & \cellcolor{blue!49}{\textcolor{black}{0.194}} & \cellcolor{blue!57}{\textcolor{black}{0.083}} \\
Bg & Texture & hor. str. & \cellcolor{blue!45}{\textcolor{black}{0.272}} & \cellcolor{blue!56}{\textcolor{black}{0.133}} & \cellcolor{blue!56}{\textcolor{black}{0.144}} & \cellcolor{blue!55}{\textcolor{black}{0.108}} \\
Bg & Texture & ver. str. & \cellcolor{blue!49}{\textcolor{black}{0.232}} & \cellcolor{blue!56}{\textcolor{black}{0.139}} & \cellcolor{blue!47}{\textcolor{black}{0.216}} & \cellcolor{blue!57}{\textcolor{black}{0.087}} \\
Bg & Texture & dots & \cellcolor{blue!45}{\textcolor{black}{0.271}} & \cellcolor{blue!55}{\textcolor{black}{0.143}} & \cellcolor{blue!49}{\textcolor{black}{0.200}} & \cellcolor{blue!55}{\textcolor{black}{0.106}} \\
Bg & Texture & diag. str. & \cellcolor{blue!44}{\textcolor{black}{0.273}} & \cellcolor{blue!55}{\textcolor{black}{0.154}} & \cellcolor{blue!50}{\textcolor{black}{0.191}} & \cellcolor{blue!53}{\textcolor{black}{0.131}} \\
Bg & Texture & con. rgs & \cellcolor{blue!43}{\textcolor{black}{0.287}} & \cellcolor{blue!54}{\textcolor{black}{0.158}} & \cellcolor{blue!40}{\textcolor{black}{0.268}} & \cellcolor{blue!57}{\textcolor{black}{0.091}} \\
Bg & Texture & bubbles & \cellcolor{blue!46}{\textcolor{black}{0.261}} & \cellcolor{blue!54}{\textcolor{black}{0.163}} & \cellcolor{blue!30}{\textcolor{black}{0.346}} & \cellcolor{blue!45}{\textcolor{black}{0.230}} \\
Obj & Texture & con. cir. & \cellcolor{blue!54}{\textcolor{black}{0.185}} & \cellcolor{blue!58}{\textcolor{black}{0.104}} & \cellcolor{blue!51}{\textcolor{black}{0.177}} & \cellcolor{blue!58}{\textcolor{black}{0.075}} \\
Obj & Texture & rad. grad. & \cellcolor{blue!53}{\textcolor{black}{0.194}} & \cellcolor{blue!56}{\textcolor{black}{0.129}} & \cellcolor{blue!46}{\textcolor{black}{0.218}} & \cellcolor{blue!55}{\textcolor{black}{0.117}} \\
Obj & Texture & lin. grad. & \cellcolor{blue!54}{\textcolor{black}{0.190}} & \cellcolor{blue!58}{\textcolor{black}{0.108}} & \cellcolor{blue!40}{\textcolor{black}{0.268}} & \cellcolor{blue!56}{\textcolor{black}{0.101}} \\
Obj & Texture & ver. str. & \cellcolor{blue!54}{\textcolor{black}{0.187}} & \cellcolor{blue!52}{\textcolor{black}{0.181}} & \cellcolor{blue!41}{\textcolor{black}{0.258}} & \cellcolor{blue!58}{\textcolor{black}{0.072}} \\
Obj & Texture & hor. str. & \cellcolor{blue!48}{\textcolor{black}{0.243}} & \cellcolor{blue!51}{\textcolor{black}{0.199}} & \cellcolor{blue!40}{\textcolor{black}{0.266}} & \cellcolor{blue!56}{\textcolor{black}{0.094}} \\
Obj & Texture & dots & \cellcolor{blue!50}{\textcolor{black}{0.219}} & \cellcolor{blue!44}{\textcolor{black}{0.298}} & \cellcolor{blue!53}{\textcolor{black}{0.166}} & \cellcolor{blue!51}{\textcolor{black}{0.166}} \\
Obj & Texture & zigzag & \cellcolor{blue!48}{\textcolor{black}{0.239}} & \cellcolor{blue!49}{\textcolor{black}{0.228}} & \cellcolor{blue!31}{\textcolor{black}{0.341}} & \cellcolor{blue!56}{\textcolor{black}{0.098}} \\
Obj & Texture & checkerboard & \cellcolor{blue!32}{\textcolor{black}{0.387}} & \cellcolor{blue!43}{\textcolor{black}{0.311}} & \cellcolor{blue!20}{\textcolor{black}{0.433}} & \cellcolor{blue!57}{\textcolor{black}{0.087}} \\
Obj & Texture & diag. str. & \cellcolor{blue!18}{\textcolor{black}{0.516}} & \cellcolor{blue!31}{\textcolor{black}{0.457}} & \cellcolor{blue!10}{\textcolor{black}{0.594}} & \cellcolor{blue!60}{\textcolor{black}{0.056}} \\
Obj & Texture & cr. hatch & \cellcolor{blue!10}{\textcolor{black}{0.679}} & \cellcolor{blue!10}{\textcolor{black}{0.873}} & \cellcolor{blue!21}{\textcolor{black}{0.419}} & \cellcolor{blue!10}{\textcolor{black}{0.798}} \\
\bottomrule
\end{tabular}
\end{table}

\begin{table}[h]
\caption{Mean Relative Count Error (lower is better) for Prompt 5 across all patterns. ``Bg'' denotes \textit{Background}, ``Obj'' denotes \textit{Object} and ''diag. str.'' denotes diagonal stripes, ''ver. str.'' denotes vertical stripes, ''hor. str.'' denotes horizontal stripes, ''con. cir.'' denotes concentric circles, ''lin. grad.'' linear gradient, ''rad. grad.'' denotes radial gradient, ''con. rgs.'' denotes concentric rings, ''cr. hatch'' denotes cross hatch categories.}
\label{tab:error_texture_prompt5}
\renewcommand{\arraystretch}{0.85} 
\setlength{\tabcolsep}{3pt}        
\centering

\begin{tabular}{@{}llp{1.5cm}cccc@{}}
\toprule
Cat. & Feat. & Pattern & Qwen7b & Qwen32b & Intern & Kimi \\
\midrule
Bg & Texture & noise & \cellcolor{blue!60}{\textcolor{black}{0.152}} & \cellcolor{blue!60}{\textcolor{black}{0.069}} & \cellcolor{blue!50}{\textcolor{black}{0.203}} & \cellcolor{blue!59}{\textcolor{black}{0.067}} \\
Bg & Texture & lin. grad. & \cellcolor{blue!56}{\textcolor{black}{0.185}} & \cellcolor{blue!57}{\textcolor{black}{0.109}} & \cellcolor{blue!60}{\textcolor{black}{0.126}} & \cellcolor{blue!58}{\textcolor{black}{0.081}} \\
Bg & Texture & rad. grad. & \cellcolor{blue!55}{\textcolor{black}{0.192}} & \cellcolor{blue!56}{\textcolor{black}{0.121}} & \cellcolor{blue!56}{\textcolor{black}{0.153}} & \cellcolor{blue!60}{\textcolor{black}{0.065}} \\
Bg & Texture & cr. hatch & \cellcolor{blue!54}{\textcolor{black}{0.211}} & \cellcolor{blue!56}{\textcolor{black}{0.111}} & \cellcolor{blue!53}{\textcolor{black}{0.175}} & \cellcolor{blue!59}{\textcolor{black}{0.071}} \\
Bg & Texture & hor. str. & \cellcolor{blue!48}{\textcolor{black}{0.270}} & \cellcolor{blue!56}{\textcolor{black}{0.114}} & \cellcolor{blue!57}{\textcolor{black}{0.143}} & \cellcolor{blue!55}{\textcolor{black}{0.120}} \\
Bg & Texture & checkerboard & \cellcolor{blue!49}{\textcolor{black}{0.252}} & \cellcolor{blue!56}{\textcolor{black}{0.123}} & \cellcolor{blue!52}{\textcolor{black}{0.188}} & \cellcolor{blue!57}{\textcolor{black}{0.098}} \\
Bg & Texture & ver. str. & \cellcolor{blue!51}{\textcolor{black}{0.234}} & \cellcolor{blue!54}{\textcolor{black}{0.146}} & \cellcolor{blue!47}{\textcolor{black}{0.222}} & \cellcolor{blue!58}{\textcolor{black}{0.087}} \\
Bg & Texture & dots & \cellcolor{blue!45}{\textcolor{black}{0.293}} & \cellcolor{blue!57}{\textcolor{black}{0.106}} & \cellcolor{blue!49}{\textcolor{black}{0.211}} & \cellcolor{blue!55}{\textcolor{black}{0.114}} \\
Bg & Texture & diag. str. & \cellcolor{blue!46}{\textcolor{black}{0.282}} & \cellcolor{blue!53}{\textcolor{black}{0.163}} & \cellcolor{blue!50}{\textcolor{black}{0.202}} & \cellcolor{blue!56}{\textcolor{black}{0.110}} \\
Bg & Texture & con. rgs & \cellcolor{blue!45}{\textcolor{black}{0.295}} & \cellcolor{blue!51}{\textcolor{black}{0.189}} & \cellcolor{blue!40}{\textcolor{black}{0.283}} & \cellcolor{blue!56}{\textcolor{black}{0.103}} \\
Bg & Texture & bubbles & \cellcolor{blue!48}{\textcolor{black}{0.269}} & \cellcolor{blue!51}{\textcolor{black}{0.187}} & \cellcolor{blue!30}{\textcolor{black}{0.360}} & \cellcolor{blue!49}{\textcolor{black}{0.191}} \\
Obj & Texture & con. cir. & \cellcolor{blue!55}{\textcolor{black}{0.198}} & \cellcolor{blue!59}{\textcolor{black}{0.082}} & \cellcolor{blue!51}{\textcolor{black}{0.190}} & \cellcolor{blue!58}{\textcolor{black}{0.081}} \\
Obj & Texture & lin. grad. & \cellcolor{blue!57}{\textcolor{black}{0.175}} & \cellcolor{blue!56}{\textcolor{black}{0.114}} & \cellcolor{blue!45}{\textcolor{black}{0.241}} & \cellcolor{blue!56}{\textcolor{black}{0.103}} \\
Obj & Texture & rad. grad. & \cellcolor{blue!54}{\textcolor{black}{0.206}} & \cellcolor{blue!55}{\textcolor{black}{0.126}} & \cellcolor{blue!50}{\textcolor{black}{0.199}} & \cellcolor{blue!55}{\textcolor{black}{0.117}} \\
Obj & Texture & ver. str. & \cellcolor{blue!53}{\textcolor{black}{0.219}} & \cellcolor{blue!50}{\textcolor{black}{0.205}} & \cellcolor{blue!45}{\textcolor{black}{0.245}} & \cellcolor{blue!58}{\textcolor{black}{0.083}} \\
Obj & Texture & hor. str. & \cellcolor{blue!46}{\textcolor{black}{0.284}} & \cellcolor{blue!50}{\textcolor{black}{0.205}} & \cellcolor{blue!42}{\textcolor{black}{0.269}} & \cellcolor{blue!58}{\textcolor{black}{0.087}} \\
Obj & Texture & zigzag & \cellcolor{blue!51}{\textcolor{black}{0.240}} & \cellcolor{blue!42}{\textcolor{black}{0.307}} & \cellcolor{blue!35}{\textcolor{black}{0.318}} & \cellcolor{blue!56}{\textcolor{black}{0.110}} \\
Obj & Texture & dots & \cellcolor{blue!50}{\textcolor{black}{0.245}} & \cellcolor{blue!34}{\textcolor{black}{0.416}} & \cellcolor{blue!52}{\textcolor{black}{0.186}} & \cellcolor{blue!49}{\textcolor{black}{0.187}} \\
Obj & Texture & checkerboard & \cellcolor{blue!35}{\textcolor{black}{0.397}} & \cellcolor{blue!40}{\textcolor{black}{0.338}} & \cellcolor{blue!23}{\textcolor{black}{0.414}} & \cellcolor{blue!57}{\textcolor{black}{0.094}} \\
Obj & Texture & diag. str. & \cellcolor{blue!21}{\textcolor{black}{0.532}} & \cellcolor{blue!29}{\textcolor{black}{0.482}} & \cellcolor{blue!10}{\textcolor{black}{0.603}} & \cellcolor{blue!58}{\textcolor{black}{0.088}} \\
Obj & Texture & cr. hatch & \cellcolor{blue!10}{\textcolor{black}{0.745}} & \cellcolor{blue!10}{\textcolor{black}{0.893}} & \cellcolor{blue!25}{\textcolor{black}{0.404}} & \cellcolor{blue!10}{\textcolor{black}{0.786}} \\
\bottomrule
\end{tabular}
\end{table}

\section{Attention on Visual Tokens}

\begin{figure*}[h]
    \centering
    \includegraphics[width=0.75\textwidth]{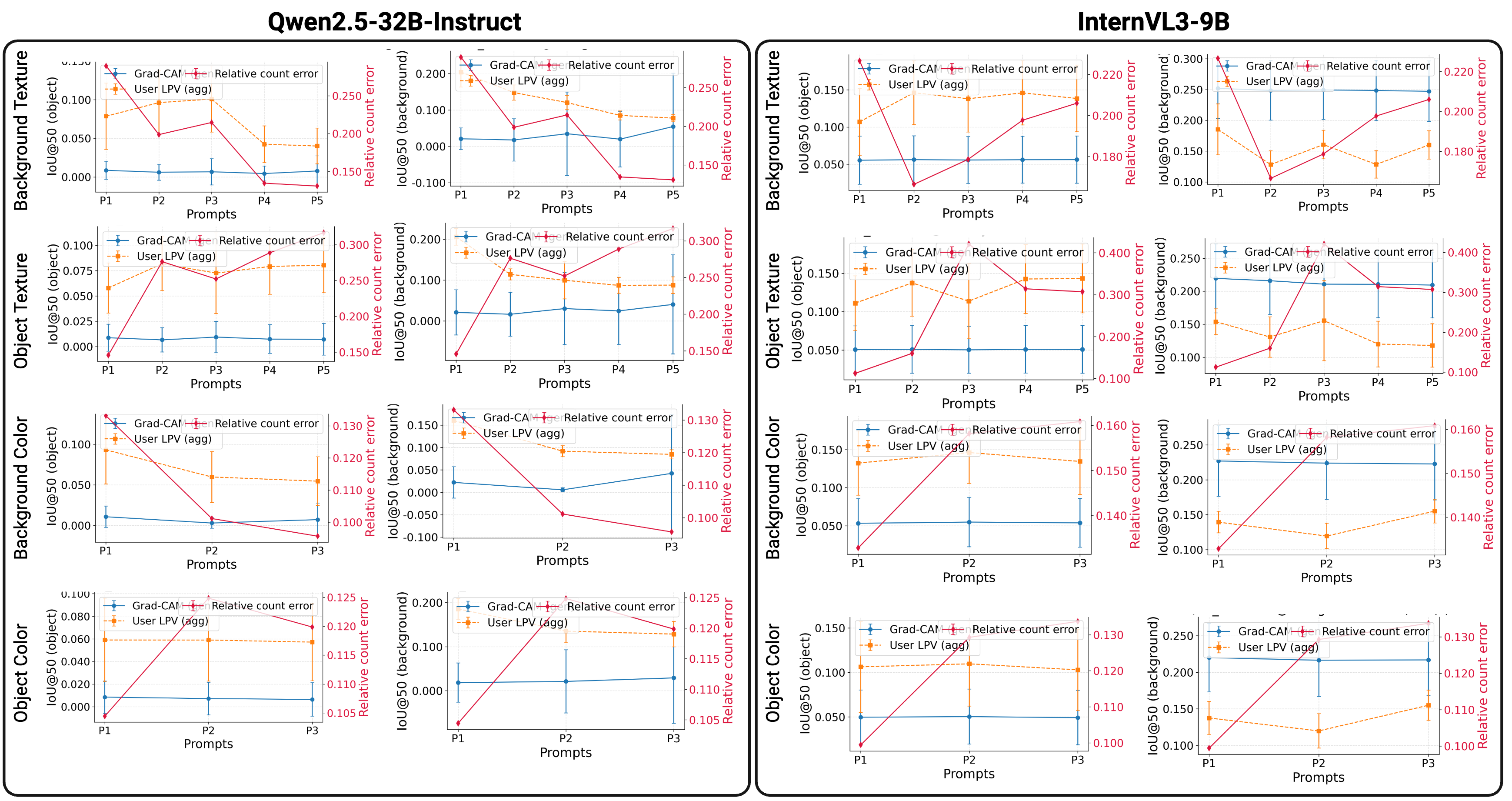}
    \caption{Visualization of the model's attention for models the Qwen2.5-32B-Instruct and InternVL3-9B}
    \label{fig:pattern_visual_attention_sup1}
\end{figure*}

\begin{figure*}[h]
    \centering
    \includegraphics[width=0.75\textwidth]{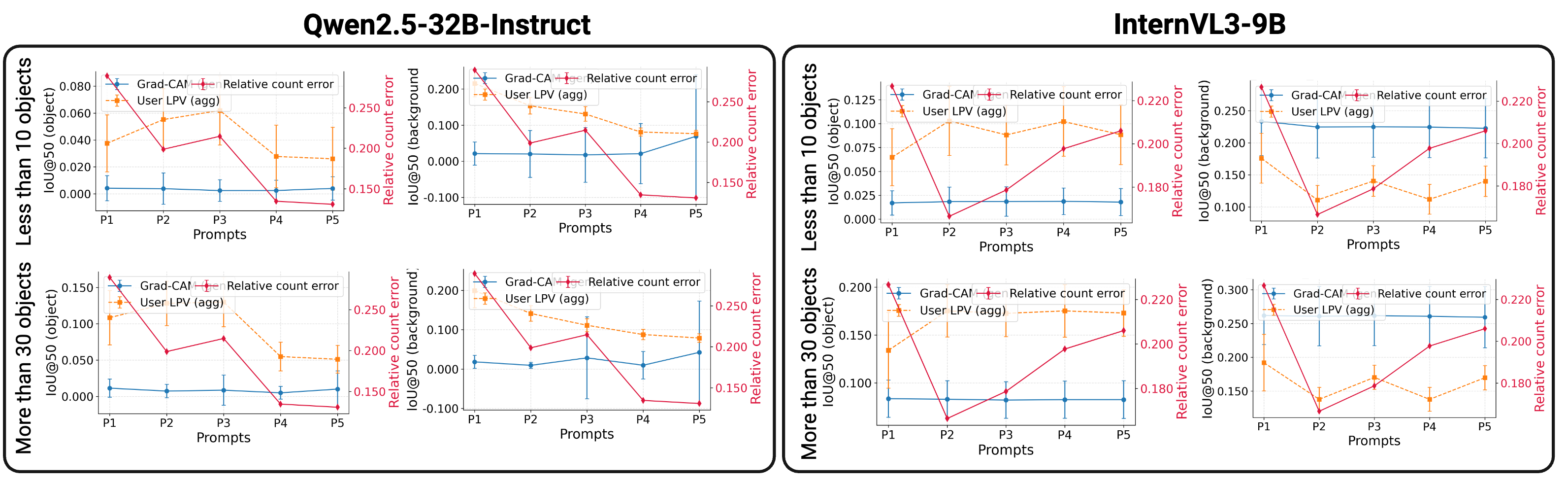}
    \caption{Visualization of the model’s attention across different background-texture patterns for images containing fewer than 10 objects or more than 30 objects for models the Qwen2.5-32B-Instruct and InternVL3-9B.}
    \label{fig:pattern_visual_attention_sup2}
\end{figure*}

Figures \ref{fig:pattern_visual_attention_sup1}-\ref{fig:pattern_visual_attention_sup2} shows the distribution of attention over vision as well as counting error across prompts for the Qwen2.5-32B-Instruct and InternVL3-9B. Across models (i.e. Qwen2.5-32B-Instruct, InternVL3-9B, Qwen2.5-7B, and Kimi-VL-A3B), we observe a consistent divide in how architectural scale influences the effect of prompt specificity. The smaller models—Qwen2.5-7B and Kimi-VL-A3B (3B active parameters)—in most cases show an initial reduction in relative count error as the prompts become more explicit, but this improvement saturates and eventually plateaus or even reverses. In contrast, the larger or more vision-specialized models—Qwen2.5-32B-Instruct and InternVL3-9B—do not exhibit this behavior. For these models, in most cases, increasing linguistic specificity does not reliably improve performance; their error remains relatively stable or fluctuates despite more detailed instructions. These findings suggest that, unlike smaller models that benefit from increased prompt granularity, higher-capacity or vision-specialized architectures may not gain additional advantage from more explicit linguistic guidance for this counting task.

\end{document}